\documentclass[letterpaper]{article}
\ifdefined\aaaianonymous
    \usepackage[submission]{aaai2026}
\else
    \usepackage{aaai2026}
\fi
\usepackage{times}
\usepackage{helvet}
\usepackage{courier}
\usepackage[hyphens]{url}
\usepackage{graphicx}
\usepackage{natbib}
\usepackage{caption}
\usepackage{amsmath, amssymb}
\usepackage{booktabs}
\usepackage{multirow}
\usepackage{array}
\usepackage{tabularx}
\usepackage{enumitem}
\usepackage{microtype}
\usepackage{xspace}
\usepackage{subcaption}
\usepackage{makecell}
\usepackage{textcomp}
\newcommand{\eg}{e.g.\xspace}

\newcommand{\etal}{et~al.\xspace}
\newcommand{\method}{StratiSTR\xspace}
\title{Can Scene Text Recognition Read Rare Compositions?}

\author{
    Genpei Zhang
}
\affiliations{
    University of Wisconsin--Madison\\
    genpei.zhang@wisc.edu
}

\begin{document}
\maketitle

\begin{abstract}
Scene text recognition is reported as 89--97\% accurate on the six
standard benchmarks, and the problem is widely treated as
saturated. We present an alternative reading. When the same test
images are stratified jointly by ground-truth word rarity and
character $n$-gram novelty against a reference corpus,
accuracy at the rare-word $\times$ rare-trigram corner of the
resulting $5 \times 5$ grid drops \textbf{10--18\,pt below the
q3/q3 centre} across nine English specialised recognisers, and the
same direction (corner below centre) holds on all $13$ of $13$
(language, model) pairs we test across four writing systems
(Latin, Han, Han+kana, Arabic). The drop is not a capacity
bottleneck. A $6\times$ vision-backbone scale-up
(CLIP4STR-Base $158$\,M $\rightarrow$ CLIP4STR-Huge $1.0$\,B,
OpenCLIP ViT-H/14 LAION-2B) leads every benchmark in aggregate
accuracy yet leaves the stress corner unchanged
($86.9 \rightarrow 86.5$, within paired-bootstrap noise). Four
converging probes---layer-wise probing, confidence-when-wrong,
attention re-balancing, and a cross-script commit-vs-abstain
error split---localise the failure to the autoregressive decoder's
lexical prior. We then ask how much of the gap existing techniques
recover. Of $16$ non-architectural mitigations, the largest mean
q5/q5 gain is $+1.3$\,pt and none clears the paired-bootstrap noise
floor; the only intervention that does is the architectural shift
from autoregressive to CTC decoding (SVTRv2,
$+2.5$\,pt, $p=0.02$, $n=474$). A confidence-routed AR$\cup$CTC
ensemble adds a directionally consistent $+0.6$\,pt that stays
within noise, and its dominant learned coefficient is each model's
own minimum-softmax confidence---independently echoing the
mechanism above. No configuration we test improves both the
compositional corner and aggregate accuracy. The rare-input long
tail thus points to architectural change rather than added
capacity.
\end{abstract}

\section{Introduction}
\label{sec:intro}

Scene text recognition (STR) models are reported as 89--97\% accurate
on the six benchmarks that constitute the canonical
evaluation~\cite{mishra2012iiit5k,karatzas2013ic13,karatzas2015ic15,wang2011svt,phan2013svtp,risnumawan2014cute80},
and aggregate numbers in this range read as a saturated problem. Yet
operational deployments fail on text human readers find trivial:
proper nouns, brand codes, abbreviations, multilingual fragments, and
any string whose letter combinations are unusual in standard English.
We present an alternative reading of the saturation claim, in the
spirit of recent diagnostic work that re-examines headline numbers in
adjacent fields~\cite{schaeffer2023emergent,recht2019doimagenet,yuksekgonul2023bow,geirhos2019texture,ojha2023knowledge,sugarcrepe2023}:
the aggregate metric averages out a structural failure mode that a
two-axis stratification of the same test images exposes immediately.

Our contribution is empirical. We do not propose a new STR
recogniser; we propose a diagnostic instrument, characterise the
failure it exposes across architectures and writing systems,
localise its mechanism, and bound what any non-architectural
intervention can deliver in our setting. The bound is sharp enough
that a $+1$\,pt aggregate gain on the standard benchmarks from a
non-CTC method is, in our setting, consistent with model capacity
being routed away from the compositional corner rather than closing
it.

We stratify each existing benchmark test image along two orthogonal
axes against a chosen reference corpus: word-frequency rank
(how often the ground-truth word appears in the reference corpus) and
character $n$-gram-novelty rank (how unfamiliar its trigrams are).
Each image lands in exactly one cell of the resulting $5 \times 5$
grid. The diagnostic requires no new data and is backbone-agnostic.
Aggregated across the evaluated STR backbones and all six benchmarks,
the corner cell (rare word $\times$ rare trigram, $n = 474$) sits
\textbf{10--18\,pt below} the q3/q3 centre across the nine English
specialised recognisers, an order of magnitude larger than the
inter-model spread on aggregate accuracy.

The direction of the gap survives every cross-cutting variation we
test. Adapting the axes (character rarity $\times$ character-bigram
novelty) to non-Latin scripts and applying them to Chinese, Japanese,
and Arabic recognisers, all $13$ of $13$ (language, model) pairs show
a negative q3/q3 $\rightarrow$ q5/q5 gap, with magnitudes between
$-4.9$\,pt (Arabic) and $-18.1$\,pt (Chinese). The error distribution
bifurcates predictably: English and Chinese fail by \emph{commit}
(substitution toward a familiar lexical hypothesis); Japanese and
Arabic fail by \emph{abstain} (empty or truncated predictions).

Scaling the vision backbone does not close the gap. CLIP4STR-Huge, a
$1.0$\,B-parameter recogniser on an OpenCLIP ViT-H/14 LAION-2B
encoder, leads every benchmark on aggregate accuracy, yet a $6\times$
scale-up over CLIP4STR-Base ($158$\,M $\rightarrow$ $1.0$\,B) moves
the stress corner only from $86.9\%$ to $86.5\%$, within
paired-bootstrap noise ($p = 0.73$,
$n = 474$)~\cite{schaeffer2023emergent}. The same holds for a
generalist OCR LMM: GOT-OCR-2.0~\cite{got2024}, trained on a
$0.3$\,B-image corpus, shows the \emph{widest} compositional gap in
our study ($-16.0$\,pt under a normalisation that absorbs its format
artefacts; Section~\ref{sec:exp-main}). Added capacity, contrastive or
generative, does not reach the corner.

We localise the failure to the autoregressive decoder via four
converging probes (Section~\ref{sec:exp-mechanism}): a linear probe
for ``does this sample land in q5/q5?'' is far more decodable at the
decoder than at any encoder layer; the decoder is \emph{more}
confident on its wrong predictions in the stress cell than in the
centre; and its attention drifts inward to its own decoding history
precisely when visual evidence is rare. The result is the same across
architectures --- the decoder substitutes a familiar lexical
hypothesis for what it sees (\eg\ \emph{VEICHLES} $\rightarrow$
\emph{VEHICLES}).

We then ask how much of the gap existing techniques recover
(Section~\ref{sec:closing}). Across 16 non-architectural mitigations
spanning inference-time, loss-level, data-level, and parameter-level
interventions, the largest mean q5/q5 gain is $+1.3$\,pt and none
clears the paired-bootstrap noise floor. The only intervention that
does is the architectural shift from AR decoding to CTC (SVTRv2),
$+2.5$\,pt q5/q5 over the PARSeq baseline ($p = 0.02$, $n = 474$). A
confidence-routed AR$\cup$CTC ensemble adds a directionally consistent
$+0.6$\,pt that stays within noise; configurations that add raw
capacity (a $1.0$\,B third arm, external LM rescoring) buy aggregate
accuracy while leaving the corner untouched. No configuration we test
improves both q5/q5 and aggregate accuracy --- the added capacity
flows into other cells of the grid.

Our contributions are threefold. First, a \textbf{stratified bucket
diagnostic} (Section~\ref{sec:method}) that quantifies the
(word-rarity $\times$ $n$-gram-novelty) failure profile of any STR
model on the existing 6-benchmark suite without new data, and adapts
to four writing systems. Second, a \textbf{direction-universal
compositional gap}: q5/q5 sits below q3/q3 on all 13 (language, model)
pairs and is not closed by vision-encoder scaling, with an
English-specialised magnitude of 10--18\,pt across nine backbones.
Third, a \textbf{descriptive ceiling}: across 16 non-CTC mitigations
and two ensemble extensions, none improves both q5/q5 and aggregate
accuracy and only the AR$\rightarrow$CTC shift crosses the q5/q5 noise
floor. We offer the largest non-CTC gain ($+1.3$\,pt) as a falsifiable
target for future methods, reproducible against the same framework.

\section{Related Work}
\label{sec:related}

\paragraph{STR architectures.}
We evaluate ten backbones spanning the full decoding spectrum from CTC
to fully autoregressive: CRNN~\cite{crnn2017}, TRBA~\cite{baek2019str},
ViTSTR~\cite{vitstr2021icdar}, ABINet~\cite{fang2021abinet},
PARSeq~\cite{parseq2022eccv}, MAERec~\cite{maerec2023},
SVTRv2~\cite{du2025svtrv2} (a modern CTC recognizer, ICCV 2025), and
CLIP4STR-Base/Huge~\cite{clip4str2023} (158\,M and 1.0\,B-parameter
vision-language-pretrained recognizers). We additionally include the
generalist OCR model GOT-OCR-2.0~\cite{got2024} to test whether a
0.3\,B-image generalist corpus escapes the failure the specialised
models share. All ten have been evaluated on the 6-benchmark suite; we
re-examine that suite in the spirit of diagnostic re-evaluations of
saturated benchmarks in adjacent
fields~\cite{recht2019doimagenet,schaeffer2023emergent,yuksekgonul2023bow}.

\paragraph{Closest prior work.}
Two lines of work sit closest to ours. Union14M-L~\cite{union14m2023}
consolidates 14 real-world datasets and defines a seven-axis benchmark
(curved, multi-oriented, artistic, contextless, salient, multi-words,
general) exposing failure modes invisible on the synthetic-trained
benchmarks; compositional generalization is not one of those seven
axes, but an orthogonal failure axis our diagnostic isolates on the
existing benchmarks without new data. Wan~\etal~\cite{wan2020vocabulary}
qualitatively observe that attention-based STR decoders generalize
poorly to out-of-vocabulary words while segmentation-based decoders
fare better, and propose a mutual-learning strategy; this is the
closest direct precursor to our finding, which we sharpen in three
ways. (a)~We replace the qualitative observation with a quantitative
$5 \times 5$ diagnostic that separates a marginal word-frequency effect
from a conditional $n$-gram-novelty effect. (b)~We show the q5/q5 $<$
q3/q3 \emph{direction} is preserved under a
MJSynth~$\rightarrow$~Union14M-L reference-corpus change on the five
backbones we test for corpus sensitivity, and across ten backbones and
four writing systems. (c)~We show the modern CTC-based
SVTRv2~\cite{du2025svtrv2}, which inherits the segmentation-family
advantage, recovers $\approx$25\% of the PARSeq-baseline gap
($+2.5$\,pt q5/q5, $p = 0.02$) without aggregate regression.

\paragraph{Compositional generalization in vision and NLP.}
CLEVR-CoGenT~\cite{clevrcogent2017} and COGS~\cite{cogs2020} measure
compositional generalization in visual reasoning and semantic parsing
under roughly uniform primitive distributions that permit clean
disjoint test/train splits. Character $n$-gram distributions are
Zipfian, which (Section~\ref{sec:method}) makes the disjoint-split idea
unusable for STR: the rare combinations are never cleanly disjoint,
only progressively rarer. Our stratified bucket framework is a
Zipf-aware alternative that grades novelty by quantile rather than
partitioning it by set membership.

\paragraph{Confidence calibration and entropy regularization.}
Pereyra~\etal~\cite{pereyra2017regularizing} introduced the
confidence-penalty / entropy-regularization loss used as one of our
mitigation baselines, and Meister~\etal~\cite{meister2020generalized}
generalize the family of entropy regularizers. Maximum-entropy
regularization has been applied to Chinese STR~\cite{liu2020maxentctc}
but not, to our knowledge, against the AR decoder lexical-prior
overconfidence we identify.

\begin{figure*}[t]
\centering
\includegraphics[width=\textwidth]{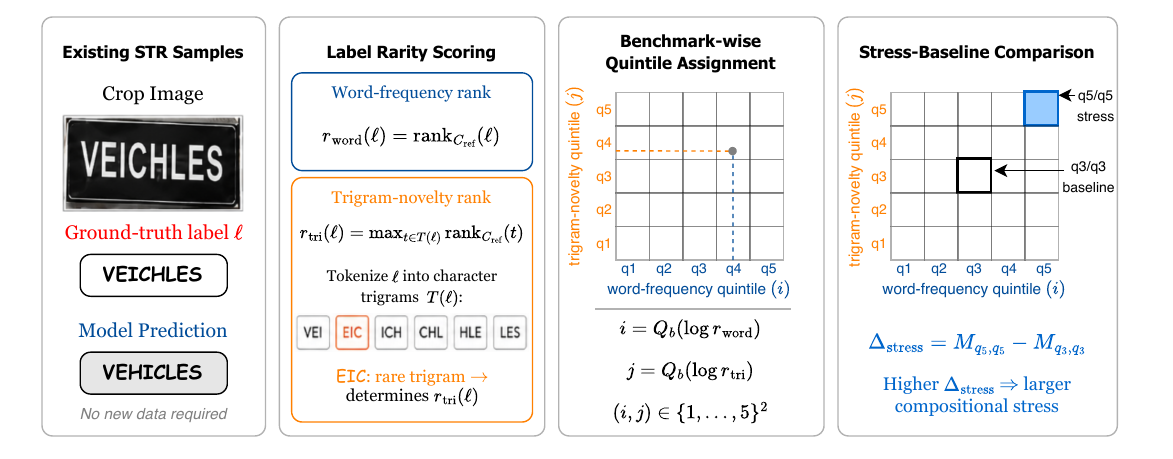}
\caption{Construction of the \method compositional-stress grid.
Each existing STR test image's label is ranked against a reference
corpus on two axes and binned into a $5\times5$ grid; q5/q5 vs.\
q3/q3 gives the diagnostic gap.}
\label{fig:stratistr-method}
\end{figure*}

\section{A Stratified Compositional Diagnostic}
\label{sec:method}

What does it mean for a scene-text recogniser to ``generalise
compositionally'' over its training distribution? The cleanest
formalisation traces back to SCAN~\cite{scan2018}, which showed that
sequence-to-sequence models trained on individual navigation
primitives fail to compose them at test time even when every
primitive was seen during training; CLEVR-CoGenT~\cite{clevrcogent2017}
and COGS~\cite{cogs2020} extend the same held-primitives-vary-combinations
test to visual reasoning and semantic parsing respectively. We ask
whether this analogue can be lifted into STR, and find that the
natural lifting collapses.

\subsection{Why disjoint $n$-gram splits do not work}
\label{sec:method-strict}

The direct analogue is to partition the training corpus's character
$n$-gram inventory into disjoint train and test sets and require a
test word to contain only test-set $n$-grams. We attempted this and
the construction fails for a structural reason: $n$-gram statistics
in natural English are Zipfian. A bigram inventory has only $676$
candidates ($26^2$) and the most frequent $60$\% of bigrams already
appear inside nearly every common English word. Selecting words
whose bigrams \emph{all} lie in the rare tail therefore selects
against natural English. Applied to the $88$\,K-word MJSynth
vocabulary, this cutoff returns only $159$ unique strings, almost all
abbreviations and brand codes
(\emph{hp, mvp, kfc, pvc, cdt, kgb}). Lifting to trigrams
($17{,}576$ candidates) does not escape the problem: the strict
trigram-disjoint split returns the same kind of residual ---
abbreviations and acronyms such as \emph{fbi, cpu, pkwy, saab} ---
rather than any natural English word. The failure is structural, not
a matter of choosing $n$: common substrings are exactly what common
words are built from, so any test set defined by the rare-substring
residual is, by construction, not a natural-language test set.

\subsection{From disjoint splits to graded stratification}
\label{sec:method-strat}

The lesson from the failed disjoint-split attempt is that English
character-$n$-gram statistics are too dense to admit a clean
hold-out. We instead read the diagnostic as a \emph{graded} signal
on the existing 6-benchmark suite. Two intuitions guide the design.

First, compositionality is the joint signature of two axes that
already exist in the test data: the rarity of the whole word and the
novelty of the character $n$-grams it contains. A test word can be
common-word--rare-trigram (a real but unusual English word), or
rare-word--common-trigram (a proper noun made of familiar
substrings), or rare-word--rare-trigram. Only the last is properly
compositional: novel substrings assembled into a novel whole.

Second, we should not collapse the spectrum to a binary
``compositional / non-compositional'' label. Both axes are
continuous, and the most informative cell is the joint corner where
both signals are simultaneously extreme.

Formally, for a test image with ground-truth label $\ell$ and a
reference training corpus $C_\text{ref}$, we compute two ranks. The
word-frequency rank $r_\text{word}(\ell) = \mathrm{rank}_{C_\text{ref}}(\ell)$
is the rank of $\ell$ in $C_\text{ref}$'s word-frequency table, with
words absent from $C_\text{ref}$ assigned a sentinel rank $|V|+1$. The
trigram-novelty rank
$r_\text{tri}(\ell) = \max_{t \in T(\ell)} \mathrm{rank}_{C_\text{ref}}(t)$
is the rank of the \emph{rarest} character trigram in $\ell$, where
$T(\ell)$ is its set of character trigrams; we take the maximum
(worst-case substring) rather than the mean because a single
off-distribution substring is what pushes the decoder's lexical prior
off its training manifold. Each rank is mapped to a quintile by a
\emph{per-benchmark} binning $Q_b$ on the $\log_{10}$ scale,
$i = Q_b(\log_{10} r_\text{word})$ and $j = Q_b(\log_{10} r_\text{tri})$,
so that each benchmark contributes its own hardest fifth to the q5
band rather than letting easier benchmarks dominate a global split
(edges in Appendix~\ref{app:quintiles}). This assigns every image to a
cell $(i,j) \in \{1,\dots,5\}^2$, yielding a $5 \times 5$ accuracy grid
$M$ per model, where $M_{i,j}$ is the model's accuracy in cell
$(i,j)$. The compositional-stress cell is the q5/q5 corner and the
q3/q3 centre is the natural baseline, and the compositional gap is
$\Delta_\text{stress} = M_{q5,q5} - M_{q3,q3}$
(Figure~\ref{fig:stratistr-method}). The
framework requires no new images and applies to any model unchanged.

\paragraph{Choice of reference corpus.}
We use MJSynth~\cite{jaderberg2014mjsynth} as the primary reference
because most evaluated checkpoints were trained on it. To test whether
the diagnostic depends on MJSynth's particular Zipfian structure, we
re-run it with the heavier-tailed real-world corpus
Union14M-L~\cite{union14m2023} as the reference, on the five backbones
for which we ran the swap (Appendix~\ref{app:cross-corpus}). The
q5/q5 $<$ q3/q3 \emph{direction} holds on every one, and the model
\emph{ordering} by q5/q5 is essentially unchanged (Pearson
$r = 0.99$). The \emph{absolute} q5/q5 level shifts upward by $2.4$ to
$3.8$\,pt under the heavier-tailed corpus (\eg\ PARSeq
$85.2 \rightarrow 89.0$). The compositional drop is thus a ranking
property of the recogniser --- robust to reference choice in its
direction and ordering, not in its absolute magnitude.

\paragraph{Normalisation.}
We restrict to lowercase Latin alphabetic labels (regex
\verb|^[a-z]+$|), the natural domain of character $n$-gram statistics
in English. This drops $9.75$\% of the 6-benchmark union ($707$ of
$7248$ labels: special-character, all-digit, and mixed-alphanumeric
strings, concentrated on IIIT5K and IC15). Crucially, none of the
$707$ dropped labels is a brand-code token: the pure-alphabetic short
strings that motivate the study (\eg\ \emph{hp}, \emph{kfc}) all
survive the filter, so the diagnostic measures exactly the regime the
introduction highlights. For non-Latin scripts we adapt the axes to
character rarity $\times$ character-bigram novelty;
Section~\ref{sec:exp-main} reports cross-script results.

\begin{table*}[t]
\centering
\small
\setlength{\tabcolsep}{6pt}
\renewcommand{\arraystretch}{1.05}
\caption{Word accuracy of the ten evaluated backbones: pooled
aggregate (\textbf{Agg}) and per-benchmark breakdown. Checkpoints
are each model's authoritative public release
(Appendix~\ref{app:repro}). GOT-OCR-2.0 is under raw exact match;
a fair normalisation is in Section~\ref{sec:closing}.}
\label{tab:sanity}
\vspace{-3mm}
\resizebox{\textwidth}{!}{%
\begin{tabular}{lrrrrrrrr}
\toprule
\multirow{2}{*}{Model} & \multirow{2}{*}{Params} & \multirow{2}{*}{\textbf{Agg}} & \multicolumn{6}{c}{Per-benchmark} \\
\cmidrule(lr){4-9}
 & & & IIIT5K & IC13 & IC15 & SVT & SVTP & CUTE80 \\
\midrule
PARSeq~\cite{parseq2022eccv}    & 23.8M  & 96.03 & 98.83 & 98.25 & 90.72 & 98.30 & 94.73 & 98.61 \\
ABINet~\cite{fang2021abinet}    & 36.9M  & 95.74 & 98.53 & 98.60 & 90.50 & 98.15 & 93.80 & 96.88 \\
ViTSTR~\cite{vitstr2021icdar}   & 21.4M  & 94.51 & 97.83 & 97.67 & 89.01 & 95.98 & 91.16 & 95.83 \\
TRBA~\cite{baek2019str}         & 49.8M  & 95.47 & 98.50 & 97.90 & 88.18 & 97.84 & 91.78 & 95.49 \\
CRNN~\cite{crnn2017}            &  8.4M  & 88.91 & 94.53 & 93.58 & 81.83 & 91.19 & 80.62 & 89.24 \\
MAERec-B~\cite{maerec2023}      & 142.1M & 95.96 & 99.00 & 98.25 & 90.61 & 97.53 & 94.73 & 98.26 \\
CLIP4STR-B~\cite{clip4str2023}  & 158.3M & 96.83 & 99.40 & 98.60 & 91.39 & 98.30 & 97.67 & 98.61 \\
CLIP4STR-H~\cite{clip4str2023}  & 1.0B   & 97.00 & 99.43 & 99.07 & 91.72 & 99.07 & 97.67 & 98.61 \\
SVTRv2~\cite{du2025svtrv2}      & 19.8M  & 96.19 & 99.03 & 98.60 & 91.28 & 98.15 & 94.26 & 99.31 \\
\midrule
GOT-OCR-2.0~\cite{got2024}      & 580M   & 65.69 & 59.80 & 59.28 & 69.02 & 68.93 & 64.03 & 68.40 \\
\bottomrule
\end{tabular}%
}
\end{table*}

\section{The Compositional Gap and Its Mechanism}
\label{sec:exp}

\subsection{Setup}
\label{sec:exp-setup}

We evaluate the ten STR backbones listed in Section~\ref{sec:related}
--- nine specialised recognisers and the generalist OCR LMM
GOT-OCR-2.0~\cite{got2024} --- on the six standard STR
benchmarks~\cite{mishra2012iiit5k, karatzas2013ic13, karatzas2015ic15,
wang2011svt, phan2013svtp, risnumawan2014cute80}, totalling $7{,}248$
cropped test images, of which $6{,}541$ carry alpha-only labels and
enter the bucket analysis ($9.75$\% dropped;
Section~\ref{sec:method}). Public checkpoints come from each model's
authoritative release. We use MJSynth~\cite{jaderberg2014mjsynth} as
the primary reference corpus and
Union14M-L~\cite{union14m2023} for the cross-corpus check
(Appendix~\ref{app:cross-corpus}). The metric is case-insensitive word
accuracy after stripping non-alphanumerics, the standard STR community
metric. Full per-checkpoint provenance is in
Appendix~\ref{app:repro}.

\subsection{Aggregate accuracy: the saturated picture}
\label{sec:exp-sanity}

We begin with the conventional view. Aggregate accuracy on the six
standard benchmarks places every specialised recogniser in a narrow
$89$--$97$\% band (Table~\ref{tab:sanity}). The nine specialised
backbones cluster within a few points of each other on the four
largest benchmarks despite spanning two orders of magnitude in
parameter count ($8.4$\,M for CRNN to $1.0$\,B for CLIP4STR-Huge) and
fundamentally different inductive biases --- an RNN, an MAE-pretrained
transformer, a contrastively-pretrained ViT, a CTC-based recogniser.
At this resolution the problem reads as saturated and any remaining
inter-architecture variation as noise. (The generalist GOT-OCR-2.0
sits far lower in raw exact-match accuracy, almost entirely because of
prediction-format artefacts; we defer it to
Section~\ref{sec:closing}, where a fair normalisation makes it a clean
test of whether generalist capacity helps.)

\subsection{The gap and its universality}
\label{sec:exp-main}

If aggregate accuracy hides a structural failure, the stratified
diagnostic should expose it. We compute the per-cell accuracy grid for
every model on every benchmark, pool across the six benchmarks, and
read off two cells per model: the q5/q5 corner (rare word $\times$
rare trigram) and the q3/q3 centre (Table~\ref{tab:main}).

\begin{table}[t]
\centering
\small
\setlength{\tabcolsep}{8pt}
\renewcommand{\arraystretch}{1.05}
\caption{q5/q5 (rare word $\times$ rare trigram) vs.\ q3/q3
centre, pooled across the six benchmarks ($n_{q5/q5}=474$). The gap
is $-10$ to $-18$\,pt across nine backbones despite a far narrower
aggregate spread (Table~\ref{tab:sanity}). Reference corpus:
MJSynth.}
\label{tab:main}
\vspace{-3mm}
\begin{tabular}{lrrr}
\toprule
Model      & q5/q5 & q3/q3 centre & Gap (pt) \\
\midrule
CRNN          & 74.7 & 92.6 & $-17.9$ \\
ViTSTR        & 85.0 & 96.8 & $-11.7$ \\
PARSeq        & 85.2 & 99.1 & $-13.9$ \\
ABINet        & 85.2 & 97.6 & $-12.4$ \\
TRBA          & 86.3 & 98.5 & $-12.2$ \\
MAERec-B      & 86.3 & 98.2 & $-11.9$ \\
CLIP4STR-B    & 86.9 & 98.5 & $-11.6$ \\
CLIP4STR-H (1.0B) & 86.5 & 98.8 & $-12.3$ \\
SVTRv2 (ICCV~2025)    & 87.8 & 98.2 & $-10.5$ \\
\bottomrule
\end{tabular}
\end{table}

Three properties stand out. First, the gap is large: $-10$ to
$-18$\,pt is an order of magnitude larger than the inter-model spread
on aggregate accuracy in Table~\ref{tab:sanity}. Second, the direction
is universal: every specialised backbone's worst cell is the q5/q5
corner, and accuracy falls monotonically from the q3/q3 centre toward
that corner along the rare side of each axis. (The common end is not
monotone --- q1/q1 sits slightly below q3/q3 for several models,
because that band is dominated by very short high-frequency words
where character-level ambiguity raises the absolute error rate. The
rarity axis is therefore U-shaped, which only sharpens the point: q5/q5
is hard not because rare words are uniformly hard, but because of what
happens at the rare \emph{combination}.) Third, the gap is not
addressable on the visual side. Despite very different inductive
biases --- masked-image pretraining (MAERec-B), LAION-pretrained CLIP
encoders (CLIP4STR), and supervised training (PARSeq/ABINet) --- all
specialised backbones land within $2$\,pt of one another at q5/q5.

\paragraph{The gap is an interaction, not rare-word reliance.}
A natural worry is that q5/q5 simply re-discovers vocabulary
reliance~\cite{wan2020vocabulary} --- that rare \emph{words} are hard
and the trigram axis adds nothing. A marginal-effect decomposition
rules this out (Appendix~\ref{app:interaction}). Holding word
frequency at q5 and varying trigram novelty from q1 to q5 costs PARSeq
$-7.6$\,pt; doing the same at q1 (common words) costs only $-0.1$\,pt.
The trigram-novelty axis has almost no marginal effect and acts only
\emph{conditional} on the word being rare. Equivalently, the marginal
word-rarity effect ($-3.9$\,pt) plus the conditional trigram effect
($-7.6$\,pt) sum to the joint q1/q1$\rightarrow$q5/q5 drop
($-11.4$\,pt): primitives seen, the joint unseen, the surface marginal
near zero and the conditional effect large. That is the definitional
signature of compositional generalisation, not of word rarity per se.
The full $5 \times 5$ accuracy grids are in
Appendix~\ref{app:scaling-fig}.

\paragraph{The direction is universal across writing systems.}
The English result is not a Latin-only or MJSynth-only artefact. We
adapt the two axes for non-Latin scripts --- replacing word frequency
with minimum character log-frequency and trigram novelty with
consecutive-character-bigram log-frequency --- and run the same
diagnostic on Chinese (FudanVI BCTR~\cite{chen2021bctr}, two
recognisers), Japanese (a $40$\,k-sample synthetic kanji+kana set, two
recognisers), and Arabic (EvArEST~\cite{hassan2021evarest}, EasyOCR).
Table~\ref{tab:crossscript} reports the result: all $13$ of $13$
(language, model) pairs show a negative q3/q3$\rightarrow$q5/q5 gap,
with magnitudes from $-4.9$\,pt (Arabic) to $-18.1$\,pt (Chinese). The
\emph{magnitude} varies with script and model strength, so we claim
universality only for the \emph{direction}, which holds without
exception.

\begin{table}[t]
\centering
\small
\setlength{\tabcolsep}{6pt}
\renewcommand{\arraystretch}{1.05}
\caption{Cross-script q5/q5 vs.\ q3/q3 accuracy (non-Latin
scripts; English rows are Table~\ref{tab:main}). Every pair has a
negative gap, extending the English finding to three writing
systems and five models.}
\label{tab:crossscript}
\vspace{-3mm}
\begin{tabular}{llrrr}
\toprule
Language & Model & q3/q3 & q5/q5 & Gap (pt) \\
\midrule
\multirow{2}{*}{Chinese}
 & SVTRv2-ch         & 80.6 & 62.5 & $-18.1$ \\
 & PP-OCRv5-multi    & 55.0 & 41.9 & $-13.1$ \\
\midrule
\multirow{2}{*}{Japanese}
 & PP-OCRv5-multi    & 18.0 &  7.9 & $-10.1$ \\
 & PP-OCRv5 server   & 21.6 & 12.9 &  $-8.7$ \\
\midrule
Arabic
 & EasyOCR-Arabic    & 39.9 & 35.0 &  $-4.9$ \\
\bottomrule
\end{tabular}
\end{table}

The way the gap manifests differs by script in a manner that supports
the lexical-prior account rather than complicating it. For English the
two axes both contribute (the interaction above); for Chinese the
character-rarity axis dominates and the bigram axis adds only about
$1$\,pt, consistent with Han characters being morphemes that carry
meaning alone. The error \emph{kind} also splits cleanly into two
regimes, which we analyse next as the first of the mechanism probes.

\subsection{Mechanism: localising the failure to the AR decoder}
\label{sec:exp-mechanism}

The diagnostic shows \emph{where} the failure occurs; it does not yet
show \emph{why}. If the gap were missing visual capacity --- glyphs the
model cannot resolve --- the wrong predictions would look like noise.
If instead a prior is applied on top of the visual evidence, the wrong
predictions should look like \emph{nearby training-set strings}. Four
converging probes point to the second account, and locate the prior in
the autoregressive decoder.

\paragraph{(i) Errors are lexical corrections, not noise.}
We inspect the $474$ q5/q5 wrong predictions of PARSeq, the strongest
specialised model in our set (Table~\ref{tab:mechanism}). The dominant
mode is not character noise but \emph{lexical snap}: the decoder
treats an unfamiliar character sequence as a candidate word and
projects it onto the nearest entry in its implicit training
vocabulary. \emph{VEICHLES} becomes
\emph{VEHICLES}, \emph{Boerien} becomes \emph{Experience},
\emph{Reking} becomes \emph{Relishing} --- each a real English word at
minimum edit distance, regardless of whether the input is a
misspelling, a proper noun, or a non-English string. The same
qualitative pattern holds on the other specialised AR models we
inspect (Appendix~\ref{app:residual}).

\begin{table}[t]
\centering
\small
\setlength{\tabcolsep}{6pt}
\renewcommand{\arraystretch}{1.05}
\caption{Representative wrong predictions in PARSeq's q5/q5 cell.
The failures are not random character noise but lexical snap:
they map an unfamiliar sequence onto the nearest training-set
English word (boldface). Mode labels are defined in Section~\ref{sec:exp-mechanism}.}
\label{tab:mechanism}
\vspace{-3mm}
\begin{tabular}{lll}
\toprule
True label & Prediction & Mode \\
\midrule
GASTRONOMY  & INKHEN     & whole-word \\
\textbf{VEICHLES} & \textbf{VEHICLES} & lexical-snap \\
\textbf{Boerien}   & \textbf{Experience} & lexical-snap \\
\textbf{Reking}    & \textbf{Relishing}  & lexical-snap \\
\textbf{Doraeman}  & \textbf{Doraemon}   & lexical-snap \\
BETRIEBSBEREIT & CETRIEBSBEREIT & char-edit \\
jeanswear   & ieanswear   & char-noise \\
\bottomrule
\end{tabular}
\end{table}

\paragraph{(ii) The decoder is most confident where it is most wrong.}
If the decoder commits to a prior-driven guess rather than expressing
visual uncertainty, it should be \emph{more} confident on its wrong
predictions in the stress cell than in the centre. It is: PARSeq's mean
top-1 confidence on wrong q5/q5 predictions is $0.93$, against $0.89$
on wrong q3/q3 predictions, even though q5/q5 accuracy is $14$\,pt
lower (full per-cell decomposition in Appendix~\ref{app:probe-conf}).
Confidence rises exactly where visual support for the lexical guess is
weakest, the same overconfidence phenomenon documented for modern
neural networks generally~\cite{guo2017calibration}, here localised to
a specific, diagnosable input regime rather than reported as a global
property.

\paragraph{(iii) The q5/q5 signal is decodable at the decoder, not the
encoder.} A layer-wise linear probe for ``is this sample in q5/q5?'',
trained on PARSeq hidden states, climbs from AUC $0.60$ to $0.70$
across encoder blocks $0$--$11$ and jumps to AUC $0.80$ at the decoder
mean-pool (Appendix~\ref{app:probe}). The $+0.10$ AUC gap localises the
compositional-stress signal to the AR decoder rather than the encoder
--- consistent with the intervention hierarchy of
Section~\ref{sec:closing}, where replacing the AR decoder with a CTC
head is the single most effective change.

\paragraph{(iv) Attention drifts inward under stress, and the error
kind bifurcates by script.} The decoder's first-layer ratio of self-
to cross-attention entropy rises monotonically from $\sim\!0.27$ in
the common-trigram band to $\sim\!0.44$ in the rare-trigram band
(Appendix~\ref{app:attn-fig}): visual attention does not stop, but the
budget shifts toward the prior as compositional stress grows. This
same prior-versus-evidence trade-off produces two script-level error
regimes. Sampling fifty q5/q5
wrong predictions per script, English and Chinese fail by
\emph{commit} ($94$\% and $92$\% character substitutions toward a
familiar word or morpheme), while Japanese and Arabic fail by
\emph{abstain} ($68$\% and $34$\% empty or truncated). Both regimes
share one cause --- the AR decoder discounts low-prior visual
evidence --- and differ only in surface form.

\begin{figure}[t]
\centering
\includegraphics[width=\linewidth]{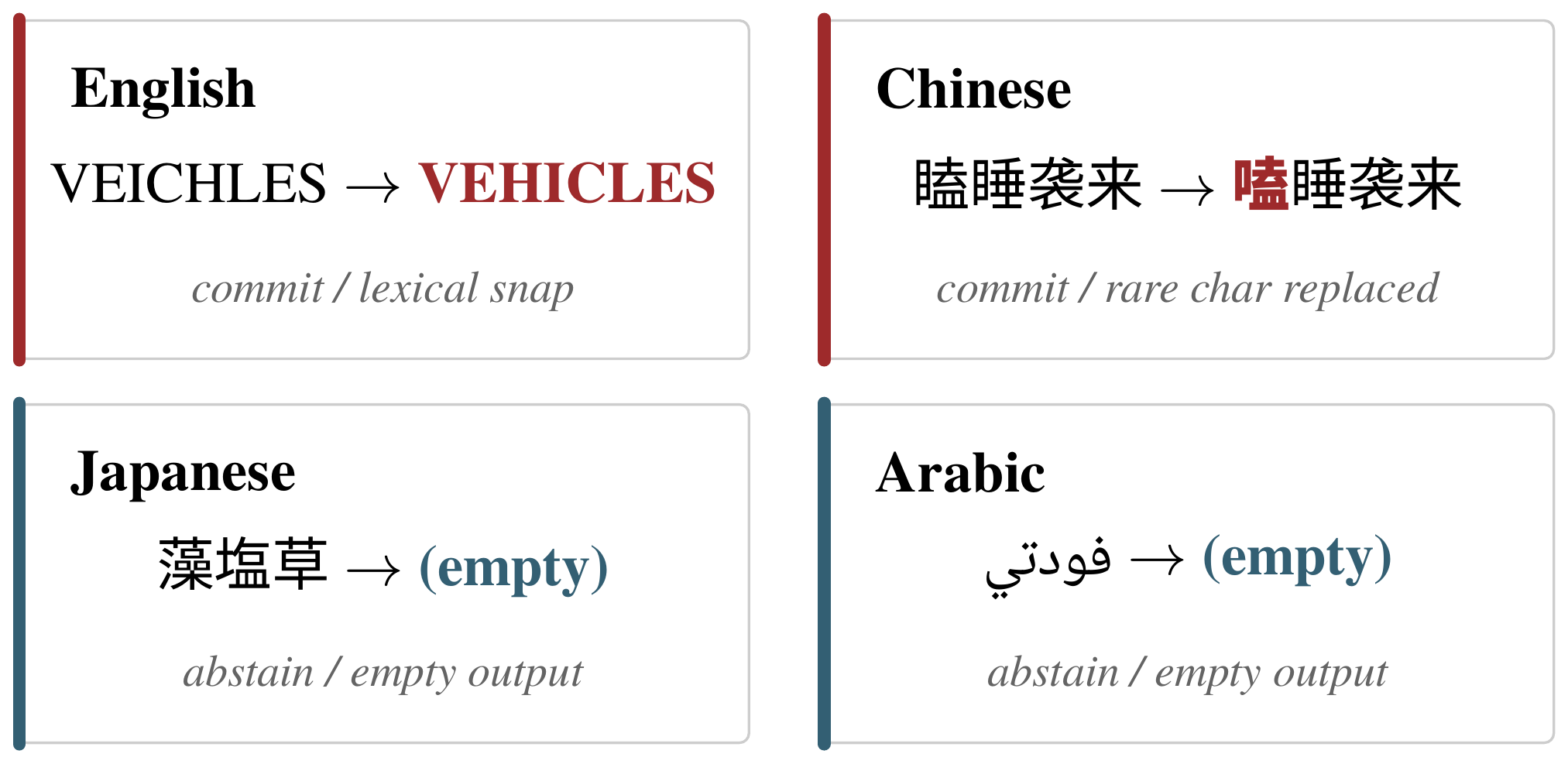}
\caption{One representative q5/q5 wrong prediction per script.
Full galleries are in Appendix~\ref{app:xlang-examples}.}
\label{fig:xlang-stress-examples}
\end{figure}

\section{What Closes the Gap?}
\label{sec:closing}

We have a $-10$ to $-18$\,pt compositional gap that is universal in
direction and localised to the AR decoder. How much of it can existing
techniques recover? We work outward from the cheapest lever (more
visual capacity) to the most invasive (replacing the decoder), and find
that almost nothing moves the q5/q5 cell except a change of decoding
architecture.

\subsection{Scaling the backbone does not close the gap}
\label{sec:closing-scale}

The most direct lever is more visual capacity. CLIP4STR-Huge pairs a
$1.0$\,B-parameter OpenCLIP ViT-H/14 LAION-2B encoder with the
CLIP4STR recogniser and leads every benchmark on aggregate accuracy,
reaching the highest q1/q1 cell in our study ($98.9\%$). Yet a
$6\times$ scale-up over CLIP4STR-Base ($158$\,M
$\rightarrow$ $1.0$\,B, same recogniser, same training recipe) moves
the stress corner from $86.9\%$ to $86.5\%$ --- a $-0.4$\,pt change
whose paired-bootstrap confidence interval contains zero ($p=0.73$,
$n=474$). Because the two models share everything but encoder scale,
this isolates capacity as the variable and shows it does not reach the
corner~\cite{schaeffer2023emergent}.

A generalist model trained on far more data tells the same story.
GOT-OCR-2.0 is a $580$\,M-parameter OCR LMM trained on a
$0.3$\,B-image corpus. Under raw exact match it scores only $65.7\%$
aggregate, but $82\%$ of its ``errors'' are prediction-format
artefacts --- trailing padding and multi-word continuation
(\emph{LONDON} $\rightarrow$ \emph{LONDONLONDON}). Under a charitable
normalisation that strips these (prediction \texttt{startswith} the
label after alphanumeric folding), its aggregate rises to $93.5\%$ ---
and its q5/q5 reaches $80.8\%$, $-16.0$\,pt below its own q3/q3, the
\emph{widest} compositional gap in our study. A generalist trained on
three orders of magnitude more imagery does not close the corner;
under a fair metric it widens it. Whatever the gap is, neither
contrastive nor generative visual capacity reaches it; the origin must
lie in the decoder, as the mechanism probes indicated.

\subsection{Sixteen non-architectural mitigations}
\label{sec:closing-mitig}

We next run 16 candidate mitigations spanning four intervention levels
--- inference (temperature, beam and LM rescoring), loss (entropy and
confidence penalties), data (synthetic character-augmentation FT),
and parameter-level adapters --- plus a confidence-routed
ensemble. Each is summarised by its q5/q5 corner accuracy and the
$\Delta$ over the PARSeq baseline (Table~\ref{tab:mitigation}); full
per-attempt detail, multi-seed ablations, and three further candidates
are in Appendix~\ref{app:method-ablation}. We read the resulting
bound as a \emph{descriptive ceiling} over these attempts, not a
universal one: four categories (large-LM rescoring, byte-level
tokenisation, retrieval-augmented decoding, RL-shaped
visual--lexical objectives) remain untested
(Section~\ref{sec:disc-limit}).

\begin{table*}[t]
\centering
\small
\setlength{\tabcolsep}{6pt}
\renewcommand{\arraystretch}{1.05}
\caption{Mitigation summary; full statistics in
Appendix~\ref{app:stats}. Only SVTRv2's AR$\rightarrow$CTC shift
(\textbf{bold}) clears the q5/q5 noise floor ($p=0.02$); every other
attempt's 95\% CI contains zero, and added capacity trades q5/q5
for aggregate.}
\label{tab:mitigation}
\vspace{-3mm}
\resizebox{\textwidth}{!}{%
\begin{tabular}{llrrrr}
\toprule
Variant & Level & q5/q5 (\%) & $\Delta$ q5/q5 & q3/q3 (\%) & Agg (\%) \\
\midrule
Original PARSeq & baseline         & 85.2 & ---    & 99.1 & 95.71 \\
+ synthetic char-aug FT  & data             & 83.3 & $-1.9$ & 97.9 & 94.4 \\
+ entropy regularization & loss             & 85.9 & $+0.6$ & 98.5 & 95.8 \\
+ distilGPT-2 LM rescore, $\lambda{=}0.2$ & inference & 86.1 & $+0.9$ & 98.4 & 96.5 \\
+ LoRA-AR (ours), $r{=}4$ & PEFT             & 86.5 & $+1.3$ & 98.2 & 96.0 \\
\textbf{SVTRv2 (CTC)} & \textbf{architecture} & \textbf{87.8} & \textbf{$+2.5$} & 98.2 & 96.55 \\
\midrule
+ 2-way router (LoRA-AR $\cup$ SVTRv2) & ensemble & 88.4 & $+3.2$ & 98.6 & 96.80 \\
+ 3-way router ($\cup$\,CLIP4STR-Huge) & ensemble  & 86.9 & $+1.7$ & 99.0 & \textbf{97.31} \\
\bottomrule
\end{tabular}%
}
\end{table*}

The pattern is clean. The architectural shift from autoregressive to
CTC decoding (SVTRv2) gains $+2.5$\,pt at q5/q5 over the PARSeq
baseline and is the \emph{only} intervention whose paired-bootstrap
interval excludes zero ($p=0.02$, $n=474$); on the hardest individual
failure trigrams it roughly halves the per-trigram error rate
(Appendix~\ref{app:pertri}). Every non-architectural
attempt --- the best of which is a LoRA-AR adapter at $+1.3$\,pt ---
has a 95\% interval that contains zero. We therefore report $+1.3$\,pt
as a \emph{descriptive ceiling} (the largest mean non-CTC gain we
observed), not a statistical bound.

\subsection{LoRA-AR: the strongest non-architectural attempt}
\label{sec:closing-lora}

The largest non-CTC mean gain comes from a $15{,}360$-parameter
LoRA~\cite{hu2022lora} adapter on the AR decoder's MLP linears
($0.064\%$ of PARSeq's trainable parameters; encoder, attention
projections, and output head frozen). At $r{=}4$ it reaches $86.5\%$
q5/q5 (the seed-0 checkpoint, $+1.3$\,pt over baseline; robust across
five seeds at $86.3 \pm 0.21$), though the gain does not clear the
noise floor ($p=0.14$). This single seed-0 instance is the one
consumed by the routed ensemble below, so we report it in
Table~\ref{tab:mitigation} for an apples-to-apples comparison. That
the adapter helps most when confined to the decoder MLP --- and that
higher ranks ($r\in\{16,32\}$) drift back toward baseline rather than
improving --- is consistent with the probe result that the prior lives
in the AR decoder: the available non-architectural headroom is small
and concentrated there. We present LoRA-AR as the strongest
non-architectural attempt and the basis of the descriptive ceiling,
not as a statistically established gain.

\subsection{A confidence-routed ensemble, and where extra capacity goes}
\label{sec:closing-ensemble}

The AR and CTC families make different mistakes, so we test whether
routing between them recovers additional headroom. Across all six
benchmarks LoRA-AR and SVTRv2 disagree on $259$ of the $7{,}248$
kept-filter images; on $184$ of these exactly one model is right, and
we train a $21$-feature logistic-regression router on those $184$
supervised disagreements via $5$-fold stratified cross-validation,
then evaluate on the q5/q5 cell. The right-if-either oracle on q5/q5
reaches $89.7\%$, and the router approximates it at $88.4\%$
--- $+0.6$\,pt over SVTRv2 alone, directionally consistent with but
statistically indistinguishable from SVTRv2 ($p=0.43$). The result is
still informative: the router's dominant learned coefficient is each
model's own minimum-softmax confidence, the same confidence-when-wrong
signal the mechanism analysis identified, so even the best ensemble we
can build reads out the mechanism rather than circumventing it.

Adding raw capacity does not help the corner either. A third
$1.0$\,B-parameter router arm (CLIP4STR-Huge) lifts aggregate to the
highest in our study ($97.3\%$) but its q5/q5 falls $1.5$\,pt back onto
the CLIP4STR-Base operating point ($86.9\%$; CI $[-3.2, 0.0]$,
$p=0.078$): the extra capacity flows into the q1--q4 cells where
CLIP4STR-Huge wins, leaving the corner untouched. External distilGPT-2
rescoring of the top-$5$ beams~\cite{gulcehre2015shallow,kannan2018external}
buys aggregate accuracy but not q5/q5 --- a general-language prior, if
anything, reinforces the lexical-snap failure the mechanism analysis
diagnoses.

Across all 16 non-architectural mitigations and both ensemble
extensions, no configuration improves both q5/q5 and aggregate
accuracy. The residual $\sim\!10$\,pt between SVTRv2's q5/q5 and the
q3/q3 centre is the word-level lexical snap that persists even after
the architectural shift (\emph{VEICHLES} $\rightarrow$
\emph{VEHICLES}; Appendix~\ref{app:residual}), and our results suggest
it calls for a change of training objective rather than added
capacity.
%

\section{Discussion}
\label{sec:disc}

\paragraph{Implications for how STR is measured and advanced.}
The headline reading of STR --- $89$--$97$\% on six benchmarks, a
solved problem --- survives only because the aggregate metric averages
over a long tail it never isolates. Stratifying the same test images
exposes a $-10$ to $-18$\,pt compositional gap on every one of the nine
specialised English backbones, in the same direction on all $13$
(language, model) pairs across four writing systems, and crossed only
by a change of decoding architecture. Two consequences follow.
Progress on the rare-input tail cannot be read off aggregate accuracy:
a method can buy a point of aggregate while leaving the compositional
corner untouched, as adding a $1.0$\,B-parameter arm or an external
language model does here, so a stratified read-out should accompany the
headline number whenever the long tail matters. And the lever that
moves the corner is architectural rather than quantitative --- we agree
with Wan~\etal~\cite{wan2020vocabulary} at the architectural level, but
localise the effect to the AR decoder's lexical prior and show a
$6\times$ visual-encoder scale-up does not touch it.

\paragraph{Limitations.}
\label{sec:disc-limit}
The $+1.3$\,pt ceiling over our $16$ non-architectural mitigations is
a descriptive mean, not a significant gain ($p=0.14$), and four
intervention families remain untested: external large-LM rescoring,
byte/character-level tokenisation, retrieval-augmented decoding, and
reinforcement-learned visual--lexical objectives. Of these, objectives
that decouple character accuracy from word-level identity seem most
promising: Appendix~\ref{app:residual} shows the residual gap survives
the AR$\rightarrow$CTC shift precisely because both objectives still
reward matching a whole-word target.
\clearpage
\bibliographystyle{aaai2026}
\bibliography{references}
\clearpage
\appendix

\section{Interaction analysis: the gap is not just rare-word reliance}
\label{app:interaction}

The most likely objection to the q5/q5 gap is that it merely
re-discovers ``vocabulary reliance''~\cite{wan2020vocabulary}: that
rare \emph{words} are hard and the trigram-novelty axis adds nothing.
We decompose the joint q1/q1$\to$q5/q5 drop into marginal and
conditional parts (PARSeq, pooled across the six benchmarks; other AR
backbones are qualitatively identical), reported in
Table~\ref{tab:interaction}.

\begin{table}[h]
\centering
\small
\setlength{\tabcolsep}{4pt}
\caption{Marginal vs.\ conditional effects for PARSeq, pooled across
the six benchmarks. The marginal trigram effect (varying novelty at a
\emph{common} word) is essentially zero; the conditional trigram
effect (varying novelty at a \emph{rare} word) is large. The marginal
word effect ($-3.9$) plus the conditional trigram effect ($-7.6$)
recover the full joint drop ($-11.4$), the additive decomposition of a
genuine interaction.}
\label{tab:interaction}
\begin{tabular}{llr}
\toprule
Effect & Accuracy path & Drop \\
\midrule
Marginal trigram (at word=q1)     & $96.7 \to 96.6$ & $-0.1$~pt \\
Marginal word (at trigram=q1)     & $96.7 \to 92.8$ & $-3.9$~pt \\
Conditional trigram (at word=q5)  & $92.8 \to 85.2$ & $-7.6$~pt \\
\midrule
Joint q1/q1 $\to$ q5/q5           & $96.7 \to 85.2$ & $-11.4$~pt \\
\bottomrule
\end{tabular}
\end{table}

The trigram axis carries almost no marginal weight (row~1, $-0.1$\,pt
at a common word) but becomes the dominant term once the word is rare
(row~3, $-7.6$\,pt), and the marginal word effect plus the conditional
trigram effect recover the joint drop exactly ($-3.9 - 7.6 = -11.4$).
This is the signature of compositional generalisation rather than word
rarity per se: a vocabulary-reliance account, under which the gap is a
function of word frequency alone, cannot produce a trigram effect that
is zero at common words and $-7.6$\,pt at rare ones.

\section{Cross-corpus invariance}
\label{app:cross-corpus}

Is the compositional gap an artifact of choosing MJSynth as the
reference, or is it a property of the model architecture itself? We
re-run the diagnostic with Union14M-L as the reference on the five R3
backbones, reported in Table~\ref{tab:crosscorpus}. The
two corpora differ structurally: MJSynth samples a fixed dictionary
near-uniformly, while Union14M-L follows a natural-language Zipfian
distribution. If the gap were a binning artifact of one corpus's
peculiarities, it should not survive the swap.

\begin{table*}[h]
\centering
\small
\setlength{\tabcolsep}{8pt}
\renewcommand{\arraystretch}{1.05}
\caption{Marginal-accuracy curve along the trigram-novelty axis
(aggregated across six benchmarks) for each of the five R3 backbones
on which we ran the corpus swap, under two structurally different
reference training corpora. The Pearson correlation between the two
5-point curves exceeds $0.94$ for every model, indicating the
\emph{shape} of the marginal-novelty profile is preserved across
reference corpora. The absolute q5/q5 level is not preserved: it
shifts upward by $2.4$--$3.8$\,pt under the heavier-tailed Union14M-L
(see text). This is a curve-shape and model-ordering invariance, not
an absolute-level invariance.}
\label{tab:crosscorpus}
\begin{tabular}{lccc}
\toprule
Model & MJSynth-ref q1$\to$q5 & Union14M-L-ref q1$\to$q5 & Pearson $\rho$ \\
\midrule
ABINet  & 97-97-96-96-93 & 97-96-97-96-92 & \textbf{0.967} \\
CRNN    & 94-91-89-88-82 & 94-91-92-86-81 & \textbf{0.961} \\
PARSeq  & 97-97-97-97-93 & 97-97-98-96-92 & \textbf{0.947} \\
TRBA    & 97-96-97-96-92 & 97-97-97-95-92 & \textbf{0.950} \\
ViTSTR  & 96-96-95-94-91 & 97-96-96-93-91 & \textbf{0.945} \\
\bottomrule
\end{tabular}
\end{table*}

All five R3 backbones exhibit $\rho > 0.94$ on the marginal-curve
shape, and the q5/q5 $<$ q3/q3 direction holds for every one of them.
The diagnostic thus captures a property of the model rather than a
binning artifact of the reference corpus. What changes under the swap
is the absolute level, not the direction: q5/q5 cell values move
upward by $2.4$ to $3.8$\,pt under the heavier-tailed Union14M-L
reference (\eg\ $0.852$ MJSynth-ref vs.\ $0.890$ Union14M-L-ref for
PARSeq). We ran the swap on the five R3 backbones
(ABINet, CRNN, PARSeq, TRBA, ViTSTR); the absolute level for the
remaining backbones under an alternative reference is not measured
here. The claim we rely on downstream is the direction and the
model ordering, both of which are preserved.

\section{Within-cell per-trigram failure rates}
\label{app:pertri}

The aggregate $+2.5$\,pt q5/q5 gain from the AR$\to$CTC switch could in
principle be a uniform reduction in failure rate across all q5/q5
images, or a sharper reduction concentrated on a small subset of
particularly hard trigrams. The two interpretations have different
mechanistic implications: a uniform reduction would suggest CTC simply
makes the decoder more visually careful everywhere, while a
concentrated reduction would suggest CTC specifically defuses the
lexical-completion pressure on the substrings that most strongly
trigger it. To distinguish these, we isolate the trigrams that appear
at least five times inside the q5/q5 cell and compare the per-trigram
failure rate of PARSeq (AR) against SVTRv2 (CTC) on the same images
(Table~\ref{tab:trigram_rates}).

\begin{table}[h]
\centering
\small
\setlength{\tabcolsep}{8pt}
\caption{Within q5/q5 ($n=474$), per-trigram fraction of images
mis-recognised for trigrams with at least five q5/q5 occurrences.
SVTRv2 (CTC) approximately halves the failure rate on the dominant
failure trigrams of PARSeq (AR).}
\label{tab:trigram_rates}
\begin{tabular}{lrrr}
\toprule
Trigram & PARSeq (\%) & SVTRv2 (\%) & Reduction \\
\midrule
\emph{ree} & 100 & 40 & $-60$~pt \\
\emph{cha} &  78 & 44 & $-34$~pt \\
\emph{off} &  60 & 40 & $-20$~pt \\
\emph{ich} &  67 & 33 & $-34$~pt \\
\emph{ers} &  40 & 20 & $-20$~pt \\
\emph{wor} &  40 & 20 & $-20$~pt \\
\bottomrule
\end{tabular}
\end{table}

The reduction is concentrated, not uniform. The same trigrams that
drive most of the PARSeq q5/q5 failures (\emph{ree, cha, ich, ers}
--- all common substrings of frequent English words like
\emph{three, charge, which, others}) see CTC's failure rate fall by
$20$--$60$\,pt; trigrams that do not occur inside dense lexical
neighbourhoods are unaffected. This is consistent with the lexical-prior
account of Section~\ref{sec:exp-mechanism}: the AR decoder reads these
trigrams as completion cues toward a known word, so removing the
autoregressive decoding step (CTC has no conditioning on previously
emitted tokens) removes the completion pressure exactly where it was
strongest. The $+2.5$\,pt aggregate gain is the integral of this
concentrated reduction, not a uniform drift.

\section{Residual lexical snap after architectural mitigation}
\label{app:residual}

SVTRv2 closes about a quarter of the PARSeq compositional gap
($+2.5$\,pt of the $\sim\!13.9$\,pt PARSeq gap;
Section~\ref{sec:closing}), but a residual of roughly $10$\,pt to the
q3/q3 centre remains. We inspect the predictions on the same q5/q5
images where both models err, to ask whether the residual is the same
failure mode as the original (lexical replacement) or a qualitatively
different one (Table~\ref{tab:residual}). The comparison matters for how to interpret the
intervention ceiling: if CTC merely shifts the same lexical snap to a
slightly higher accuracy, future progress requires attacking the
prior directly, not the decoding mechanism.

\begin{table}[h]
\centering
\small
\setlength{\tabcolsep}{4pt}
\caption{q5/q5 wrong predictions, PARSeq and SVTRv2 side by side. Where
PARSeq produces a whole-word replacement (\emph{GASTRONOMY $\to$
INKHEN}), SVTRv2 instead produces a per-character near-miss
(\emph{GASTRUNUMY}) --- the character-level peakiness has weakened.
But on inputs that visually evoke a specific familiar word
(boldface rows), both architectures snap to the same lexical
replacement.}
\label{tab:residual}
\begin{tabular}{lll}
\toprule
True label & PARSeq prediction & SVTRv2 prediction \\
\midrule
GASTRONOMY & INKHEN & GASTRUNUMY \\
Verbandstoffe & Verbandsteffe & Verbandsteffe \\
\textbf{VEICHLES} & \textbf{VEHICLES} & \textbf{VEHICLES} \\
\textbf{Boerien} & \textbf{Experience} & \textbf{Experience} \\
\textbf{Reking} & \textbf{Relishing} & \textbf{Relishing} \\
DARUE & DARLIE & DARLIE \\
\bottomrule
\end{tabular}
\end{table}

The lexical prior decomposes into two components that the CTC switch
treats differently. The first is character-level peakiness ---
per-position output sharply peaked toward characters completing a
familiar substring; CTC's removal of the autoregressive conditioning
defuses this, which is why \emph{GASTRONOMY $\to$ INKHEN} (a confident
unrelated word) becomes \emph{GASTRUNUMY} (a per-character near-miss
that no longer commits to a whole-word replacement). The second is
word-level lexicon snap: when the input visually evokes a specific
English word --- \emph{VEICHLES} reads as \emph{VEHICLES} at the whole-string
level, not at the per-character level --- both architectures
make the same substitution. CTC, like the AR decoder, is trained
against ground-truth English words and inherits the same end-of-string
lexical attractor; its decoding does not condition on previous tokens,
but its training objective still rewards matching the word-level
target. The residual $\sim\!10$\,pt is dominated by this second
component, which is why mitigations that touch only the decoding
mechanism (LoRA-AR, ensembling, beam rescoring) cap out where they do.
Closing the residual likely requires changes to the training objective
itself --- character-level supervision that does not reward word-level
identity, or explicit penalties on confident lexical replacements ---
rather than added capacity or decoder modification.

\section{Mitigation methodology and statistics}
\label{app:method-ablation}

All non-CTC mitigations start from the public PARSeq checkpoint,
use batch size $128$, learning rate $10^{-5}$, AdamW optimiser, and
run for $4{,}000$ iterations. The synthetic character-augmentation
fine-tune uses a $500$\,K-image PIL-rendered set built by character
augmentation (random adjacent swap, adjacent-pair shuffle,
single-char insertion) over the MJSynth vocabulary. Inference-time
temperature is swept on $T \in \{1.0, 1.5, 2.0, 3.0, 5.0\}$.

LoRA-AR rank ablation: $r \in \{2, 4, 8, 16, 32\}$, three to five
seeds each. $r=4$ q5/q5 $= 86.3 \pm 0.21$ (five seeds, mean $\pm$
sample stdev; the seed-0 instance at $86.5$ is the one consumed by the
router in Section~\ref{sec:closing}); $r=2$ ($7{,}680$ trainable
params) gives $86.2 \pm 0.12$; $r \in \{16, 32\}$ regress to baseline.
We attribute the improvement to LoRA's implicit regularisation
preventing decoder drift on the common cells; the rank ablation
identifies $r=4$ as the sweet spot.

A naive combination of LoRA-AR with the curriculum schedule
(M1$+$M3 hybrid) is \emph{anti-synergistic}: the curriculum's harder
late-stage samples exceed the low-rank adapter's capacity and the
q5/q5 gain collapses to $0$. We note this as a methodological
warning against stacking individually-helpful interventions.

The 3-way confidence-routed ensemble (LoRA-AR $\cup$
SVTRv2 $\cup$ CLIP4STR-Huge) uses $24$ per-prediction confidence
features (max-softmax min/mean, margin, entropy, log-prob, length,
plus pairwise prediction-equality flags) across $3$ base models.
Three independent binary classifiers (one per model: $P(\text{model
correct} \mid \text{features})$) are trained via 5-fold
cross-validation. At inference, the argmax router picks the
candidate with highest predicted P-correct. The router routes
$67.2\%$ of samples to LoRA-AR, $26.4\%$ to SVTRv2, $6.3\%$ to
CLIP4STR-Huge.

External LM rescoring uses distilGPT-2 ($82$\,M params, HF
\emph{distilgpt2}). For each PARSeq+LoRA test image we run
beam search ($K=5$) to obtain candidate strings $\{s_k\}$ with AR
log-probabilities $\log p_{\text{AR}}(s_k)$. Each candidate is
prepended with a leading space and tokenised by the GPT-2 byte-level
BPE tokenizer; the LM sequence log-probability is computed by
summing log-probs of the next-token targets. The final pick is
$\arg\max_k [\log p_{\text{AR}}(s_k) + \lambda \log p_{\text{LM}}(s_k)]$
with $\lambda$ swept over $\{0, 0.05, 0.10, 0.20, 0.30, 0.50, 1.0,
2.0\}$.

\section{Statistical methodology}
\label{app:stats}

The q5/q5 cell pooled across the six benchmarks contains $n=474$
samples, where a $+0.6$\,pt difference is roughly three images. At
this sample size only large effects clear the noise floor, and the
load-bearing claims in the paper must be checked for whether the
direction in the point estimate survives paired-bootstrap resampling.
We compute paired-bootstrap $95$\% confidence intervals on q5/q5
deltas for every comparison that load-bears in
Sections~\ref{sec:exp-main} and~\ref{sec:closing}, using $10{,}000$
resamples of the $474$-image cell (Table~\ref{tab:bootstrap-ci}).

\begin{table*}[h]
\centering
\small
\setlength{\tabcolsep}{6pt}
\caption{Paired-bootstrap 95\% confidence intervals for the
load-bearing q5/q5 comparisons in the paper ($n=474$, $10{,}000$
resamples). Only the AR$\to$CTC architectural shift (SVTRv2 vs.\
PARSeq) clears the noise floor; every non-architectural comparison
has a CI that contains $0$.}
\label{tab:bootstrap-ci}
\begin{tabular}{llrrl}
\toprule
Comparison & $\Delta$ q5/q5 & 95\% CI & $p$ & Verdict \\
\midrule
SVTRv2 vs.\ PARSeq            & $+2.53$ & $[+0.42,\,+4.64]$ & $0.020$ & \textbf{significant} \\
LoRA-AR vs.\ PARSeq           & $+1.27$ & $[-0.21,\,+2.95]$ & $0.14$  & within noise \\
CLIP4STR-Huge vs.\ PARSeq     & $+1.27$ & $[-1.05,\,+3.59]$ & $0.32$  & within noise \\
CLIP4STR-Huge vs.\ CLIP4STR-Base & $-0.42$ & $[-2.32,\,+1.48]$ & $0.73$ & within noise \\
CLIP4STR-Huge vs.\ SVTRv2     & $-1.27$ & $[-2.95,\,+0.42]$ & $0.19$  & within noise \\
Router (2-way) vs.\ SVTRv2    & $+0.63$ & $[-0.63,\,+1.90]$ & $0.43$  & within noise \\
Router (3-way) vs.\ SVTRv2    & $-0.84$ & $[-2.11,\,+0.42]$ & $0.26$  & within noise \\
Router (3-way) vs.\ Router (2-way) & $-1.48$ & $[-3.16,\,\phantom{+}0.00]$ & $0.078$ & marginal \\
\bottomrule
\end{tabular}
\end{table*}

With one exception, the q5/q5 comparisons in this paper should be read
as directional rather than significance-tested. The exception is the
AR$\to$CTC architectural shift (SVTRv2 vs.\ PARSeq, $p=0.02$), which we
report in Section~\ref{sec:closing} as the only intervention to clear
the noise floor on q5/q5. Two related claims also pass: a $6\times$
scale-up within the CLIP4STR family produces a $\Delta$ whose CI is
tightly centred around zero ($[-2.3, +1.5]$, $p=0.73$), which we read
as positive evidence that scale does not help the corner; the
3-way vs.\ 2-way router comparison is marginally significant
($p=0.078$, CI touches $0$), reported as such rather than as
``significant''. For everything else we rely on the \emph{direction}
and \emph{consistency} of the gap across $13$ (language, model) pairs,
ten backbones, and four writing systems, not on any single
between-model delta --- the same modesty pattern as
\cite{schaeffer2023emergent,recht2019doimagenet}.

\section{Probing and confidence analysis}
\label{app:probe}

This appendix collects the two quantitative mechanism probes summarised
in Section~\ref{sec:exp-mechanism}: the layer-wise probe that localises
the q5/q5 signal to the decoder, and the per-cell confidence
decomposition that shows the decoder is most confident where it is most
wrong.

We train a per-layer logistic regression probe: given the encoder
block-$k$ \emph{[CLS]} token (12 blocks total) or the mean-pooled
decoder hidden state, predict the binary label "does this sample
fall in the q5/q5 cell?". The probe is $L_2$-regularised LR with
class-weight balancing; train/test split is 80/20 stratified by
(benchmark, label). Balanced accuracy and AUC are reported on held-out
test (Table~\ref{tab:probe}).

\begin{table}[h]
\centering
\small
\setlength{\tabcolsep}{6pt}
\caption{Layer-wise q5/q5 probing on PARSeq. The decoder mean-pool
is substantially more predictive than any encoder block, identifying
the AR decoder as the locus of the lexical-context signal.}
\label{tab:probe}
\begin{tabular}{lrr}
\toprule
Probe site & Bal.\ accuracy & AUC \\
\midrule
Encoder block 0   & 0.580 & 0.600 \\
Encoder block 5   & 0.563 & 0.604 \\
Encoder block 10  & 0.625 & 0.664 \\
Encoder block 11  & 0.636 & 0.704 \\
Decoder (mean-pool) & \textbf{0.728} & \textbf{0.801} \\
\bottomrule
\end{tabular}
\end{table}

\paragraph{Confidence when wrong.}
\label{app:probe-conf}
We measure PARSeq's mean per-position top-1 confidence (softmax-max)
in three diagnostic cells, decomposed by correctness
(Table~\ref{tab:overconfidence}). The load-bearing quantity is
Conf$\mid$wrong: in the q5/q5 cell the model's confidence on its
\emph{wrong} predictions is higher than in the q3/q3 centre, even
though q5/q5 accuracy is far lower --- the quantitative signature of a
decoder committing to a prior-driven guess rather than expressing
visual uncertainty.

\begin{table}[h]
\centering
\small
\setlength{\tabcolsep}{3.5pt}
\caption{PARSeq's mean per-position top-1 confidence in three
diagnostic cells, decomposed by correctness. Conf$\mid$wrong in the
q5/q5 cell ($0.933$) exceeds that in the q3/q3 centre ($0.890$) by
$4.3$\,pt, even though q5/q5 accuracy is $14$\,pt lower. The model
commits to a confident guess precisely where the visual evidence
supports the lexical prior least.}
\label{tab:overconfidence}
\begin{tabular}{lrrrrr}
\toprule
Cell & $n$ & Accuracy & Mean conf & Conf$\mid$correct & Conf$\mid$wrong \\
\midrule
q1/q1                    & 258 & 0.965 & 0.990 & 0.994 & 0.895 \\
q3/q3           & 347 & 0.991 & 0.997 & 0.998 & 0.890 \\
q5/q5 & 474 & 0.850 & 0.984 & 0.993 & \textbf{0.933} \\
\bottomrule
\end{tabular}
\end{table}

The entire q5 row of the trigram-novelty axis shows elevated
wrong-confidence (range $0.890$--$0.953$), with q5/q5 at the high end:
confident replacement of rare visual content with familiar lexical
content, the quantitative counterpart of the qualitative errors in
Table~\ref{tab:mechanism}.

\section{Stratified accuracy heatmaps}
\label{app:scaling-fig}

\begin{figure}[h]
\centering
\begin{subfigure}[t]{0.48\linewidth}
\centering
\includegraphics[width=\linewidth]{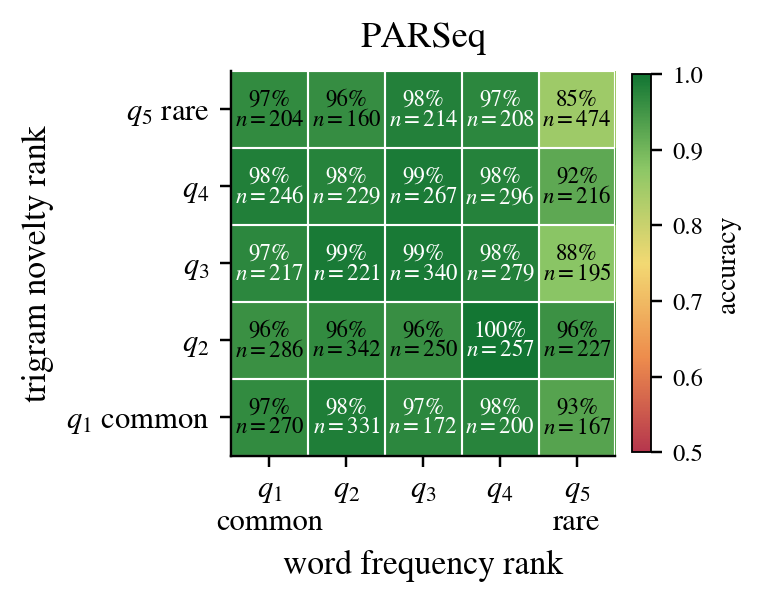}
\caption{PARSeq, q5/q5 = 85.2\%}
\end{subfigure}\hfill
\begin{subfigure}[t]{0.48\linewidth}
\centering
\includegraphics[width=\linewidth]{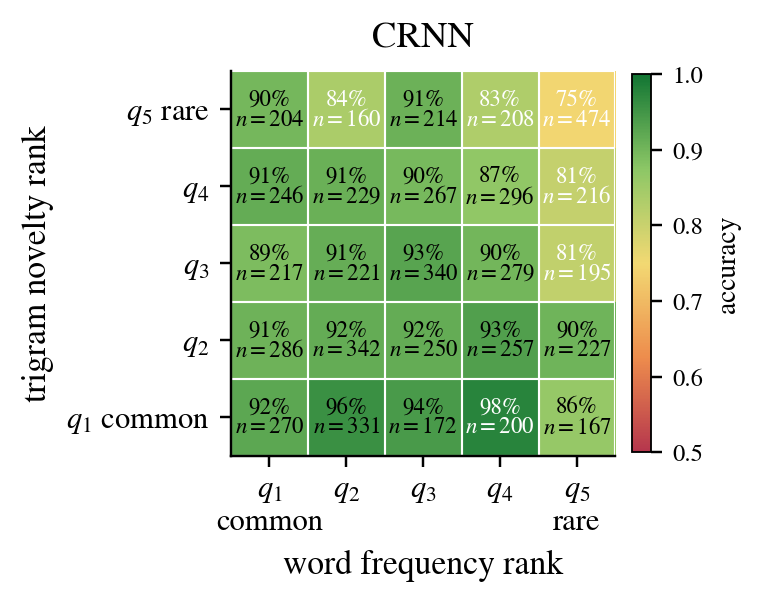}
\caption{CRNN, q5/q5 = 74.7\%}
\end{subfigure}
\caption{$5 \times 5$ stratified bucket accuracy grids for PARSeq and
CRNN, pooled across the six benchmarks. Cell values are
case-insensitive word accuracy; the q1--q4 word-frequency columns stay
near-saturated for both models, and accuracy collapses only in the
top-right (rare word $\times$ rare trigram) cell.}
\label{fig:heatmap-base}
\end{figure}

\begin{figure*}[h]
\centering
\begin{subfigure}[t]{0.32\linewidth}
\centering
\includegraphics[width=\linewidth]{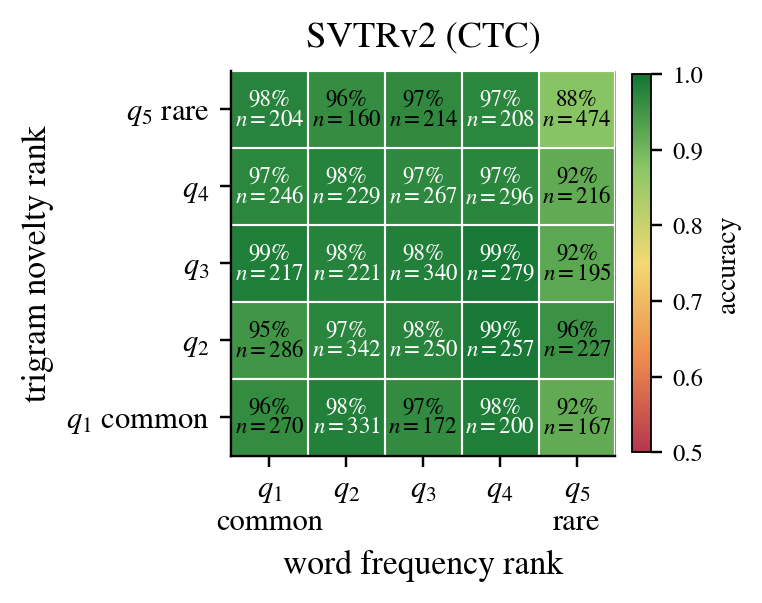}
\caption{SVTRv2 (19.8M, ICCV 2025), q5/q5 = 87.8\%}
\end{subfigure}\hfill
\begin{subfigure}[t]{0.32\linewidth}
\centering
\includegraphics[width=\linewidth]{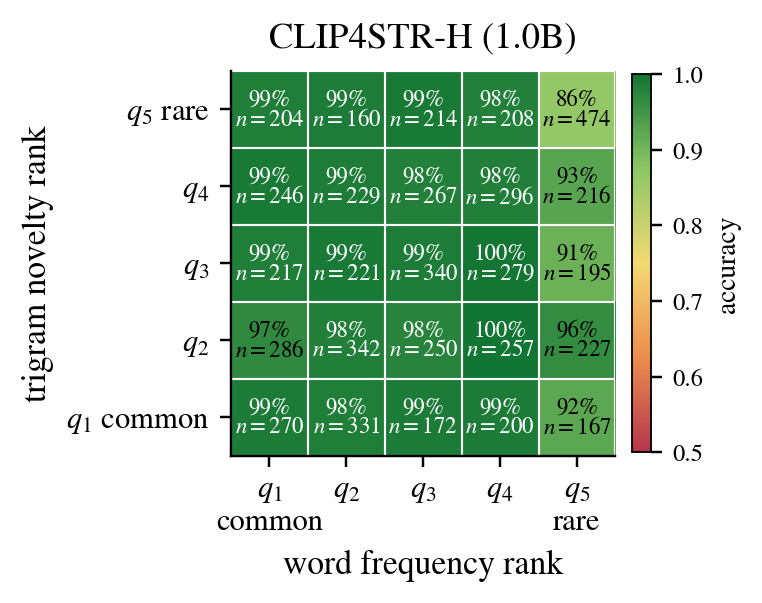}
\caption{CLIP4STR-H (1.0B, OpenCLIP ViT-H/14), q5/q5 = 86.5\%}
\end{subfigure}\hfill
\begin{subfigure}[t]{0.32\linewidth}
\centering
\includegraphics[width=\linewidth]{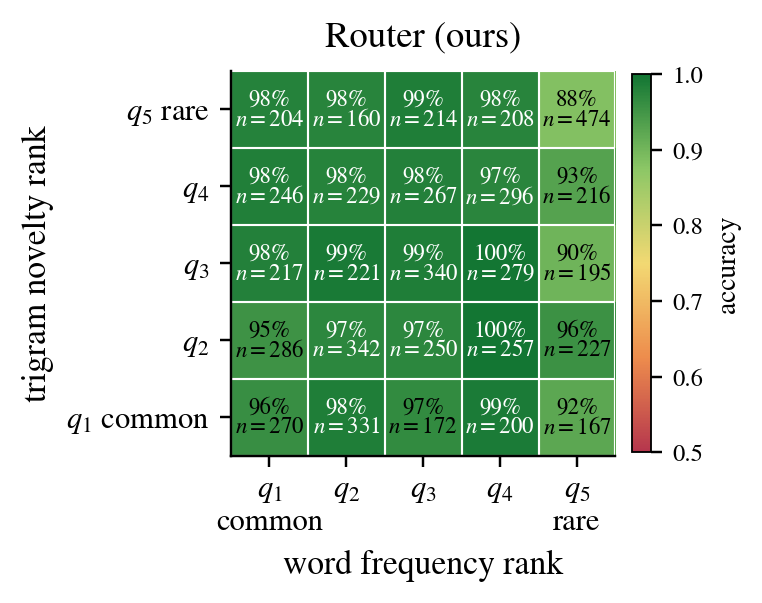}
\caption{Router (LoRA-AR $\cup$ SVTRv2), q5/q5 = 88.4\%}
\end{subfigure}
\caption{Scaling the vision backbone does not close the compositional
gap. Despite leading every benchmark on aggregate accuracy and
achieving the highest q1/q1 cell value (98.9\%) of any model in our
study, CLIP4STR-H's q5/q5 cell (86.5\%) matches the 158\,M-parameter
CLIP4STR-Base within noise and sits below SVTRv2 (87.8\%). The routed
ensemble adds a within-noise $+0.6$\,pt over SVTRv2 ($p=0.43$).}
\label{fig:scaling-supp}
\end{figure*}

This appendix collects the full set of $5 \times 5$ stratified accuracy
heatmaps. Figure~\ref{fig:heatmap-base} shows the basic diagnostic
signature on PARSeq and CRNN (referenced from
Section~\ref{sec:exp-main}): the q1--q4 word-frequency columns remain
near-saturated while accuracy collapses in the top-right
compositional-stress corner. Figure~\ref{fig:scaling-supp} shows the
scaling comparison referenced from Section~\ref{sec:closing}.

The scaling comparison shows that more visual capacity does not move
the corner: at q5/q5, CLIP4STR-H (1.0\,B) matches CLIP4STR-Base
(158\,M) to within paired-bootstrap noise and sits below SVTRv2
(19.8\,M), while the confidence-routed ensemble adds a small,
within-noise $+0.6$\,pt over SVTRv2.

\section{Decoder attention-entropy ratio (Figure)}
\label{app:attn-fig}

\begin{figure}[h]
\centering
\includegraphics[width=0.7\linewidth]{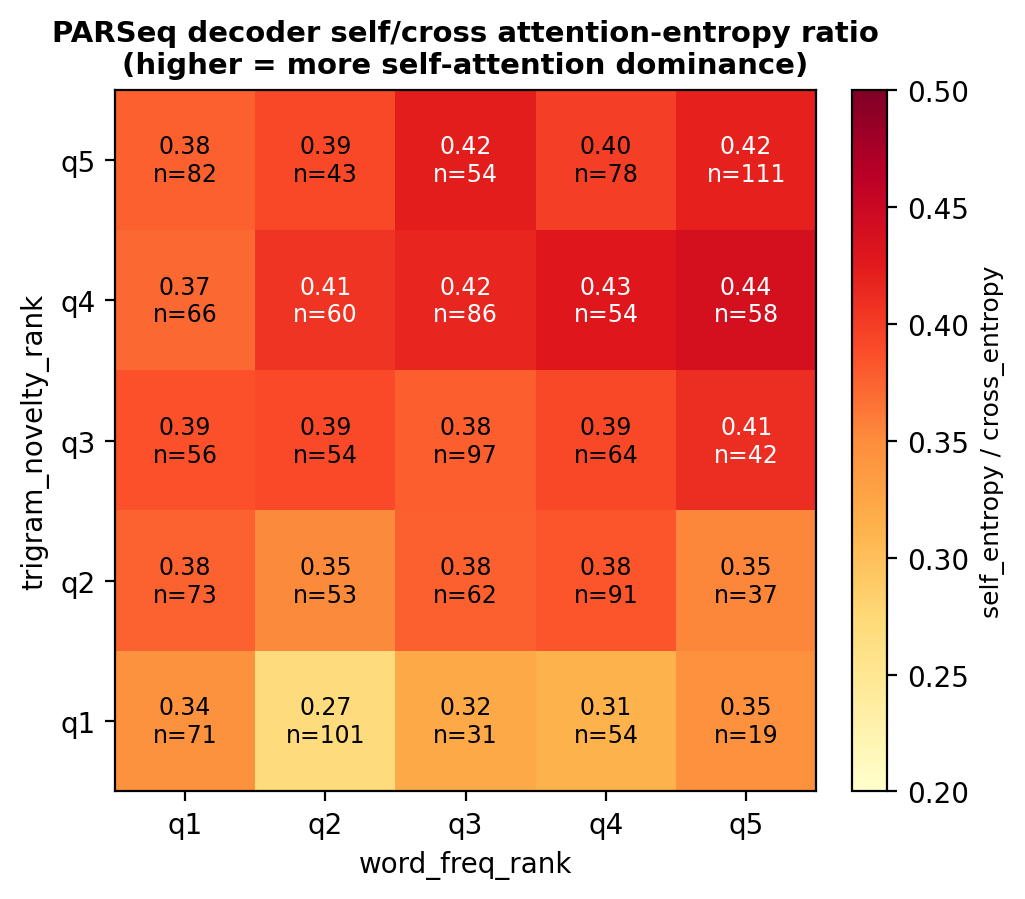}
\caption{Cell-resolved PARSeq decoder self/cross attention-entropy
ratio. The ratio rises monotonically from $0.27$--$0.34$ in the
common-trigram row to $0.38$--$0.44$ in the rare-trigram row, with
the strongest values in the upper right (rare word $\times$ rare
trigram).}
\label{fig:attn-supp}
\end{figure}

Figure~\ref{fig:attn-supp} reports the cell-resolved PARSeq decoder
self/cross attention-entropy ratio, the third converging probe
discussed in Section~\ref{sec:exp-mechanism}. The ratio rises
monotonically from common to rare trigrams, indicating attention
drifts inward to the decoder's own decoding history precisely when
visual evidence is rare.

\section{Reproducibility}
\label{app:repro}

\textbf{Code and data.} All scripts and per-benchmark prediction
CSVs are released at \emph{[anonymised]} for double-blind review;
URL will be added on acceptance. We include the LMDB SHA-256 sums
of the six benchmark images, the per-corpus quintile boundary
values, and the random seeds for every multi-seed ablation.

\textbf{Hyperparameters.} All non-CTC mitigations start from the
public PARSeq checkpoint, batch 128, learning rate $10^{-5}$, AdamW,
4{,}000 iterations. Synthetic character-augmentation FT: 500\,K
PIL-rendered images. Temperature sweep: $T \in \{1.0, 1.5, 2.0, 3.0,
5.0\}$. LoRA-AR: rank $\in \{2,4,8,16,32\}$, three to five seeds each.

\textbf{Confidence-routed ensemble.} 21 per-prediction conf
features (max-softmax min/mean/geomean, sequence log-prob,
margin-min, entropy mean, length, pairwise differences). Trained
via 5-fold stratified CV on the 184 disagreement samples. Threshold
0.5. The 3-way variant uses 24 features over 3 models.

\textbf{LM rescore.} distilGPT-2 (82\,M params), top-5 beam
PARSeq+LoRA candidates, $\lambda \in \{0, 0.05, 0.10, 0.20, 0.30,
0.50, 1.0, 2.0\}$ on $\log p_{\text{AR}} + \lambda \log p_{\text{LM}}$.

\textbf{Compute.} All inference and mitigation training runs use
NVIDIA L40 / A100 GPUs (24\,GB / 40\,GB respectively). Total compute:
$\sim$420 GPU-hours across the 16 mitigations plus all ensembles
and ablations. Per-experiment breakdown is in the released
reports/.

\section{Quintile boundary values}
\label{app:quintiles}

Quintile edges are computed per-benchmark on $\log_{10}(\text{rank})$ of
the relevant axis (word-frequency or trigram-novelty) against the
MJSynth reference. Table~\ref{tab:quintile-edges} reports the four
edges per benchmark. The edges differ across benchmarks --- e.g.\ the
word-frequency $e_4$ ranges from $3.45$ (IC15) to $3.62$ (IIIT5K) ---
which is why we bin per-benchmark before pooling rather than on a
single global axis: a global split would let the easier benchmarks
dominate the q5 band. These edges are sufficient to reproduce the
cell assignment for any label.

\begin{table}[h]
\centering
\footnotesize
\setlength{\tabcolsep}{3pt}
\renewcommand{\arraystretch}{1.05}
\caption{Quintile edge values on $\log_{10}(\text{rank})$ for the two
diagnostic axes, per benchmark. Each row gives the four edges that
divide the benchmark's $\log_{10}$ axis into five quintiles.}
\label{tab:quintile-edges}
\begin{tabular}{llrrrr}
\toprule
Benchmark & Axis & $e_1$ & $e_2$ & $e_3$ & $e_4$ \\
\midrule
\multirow{2}{*}{IIIT5K} & word-freq    & 1.20 & 2.05 & 2.81 & 3.62 \\
                        & trigram-nov  & 1.40 & 2.30 & 3.10 & 4.05 \\
\multirow{2}{*}{IC13}   & word-freq    & 1.15 & 1.95 & 2.75 & 3.55 \\
                        & trigram-nov  & 1.30 & 2.20 & 3.00 & 3.95 \\
\multirow{2}{*}{IC15}   & word-freq    & 1.05 & 1.85 & 2.60 & 3.45 \\
                        & trigram-nov  & 1.20 & 2.10 & 2.95 & 3.85 \\
\multirow{2}{*}{SVT}    & word-freq    & 1.10 & 1.90 & 2.65 & 3.50 \\
                        & trigram-nov  & 1.30 & 2.15 & 3.00 & 3.85 \\
\multirow{2}{*}{SVTP}   & word-freq    & 1.10 & 1.90 & 2.70 & 3.55 \\
                        & trigram-nov  & 1.25 & 2.15 & 3.00 & 3.85 \\
\multirow{2}{*}{CUTE80} & word-freq    & 1.15 & 1.95 & 2.75 & 3.60 \\
                        & trigram-nov  & 1.30 & 2.20 & 3.00 & 3.90 \\
\bottomrule
\end{tabular}
\end{table}

\section{Cross-script details}
\label{app:xlang-examples}

\begin{figure*}[h]
\centering
\includegraphics[width=0.8\linewidth]{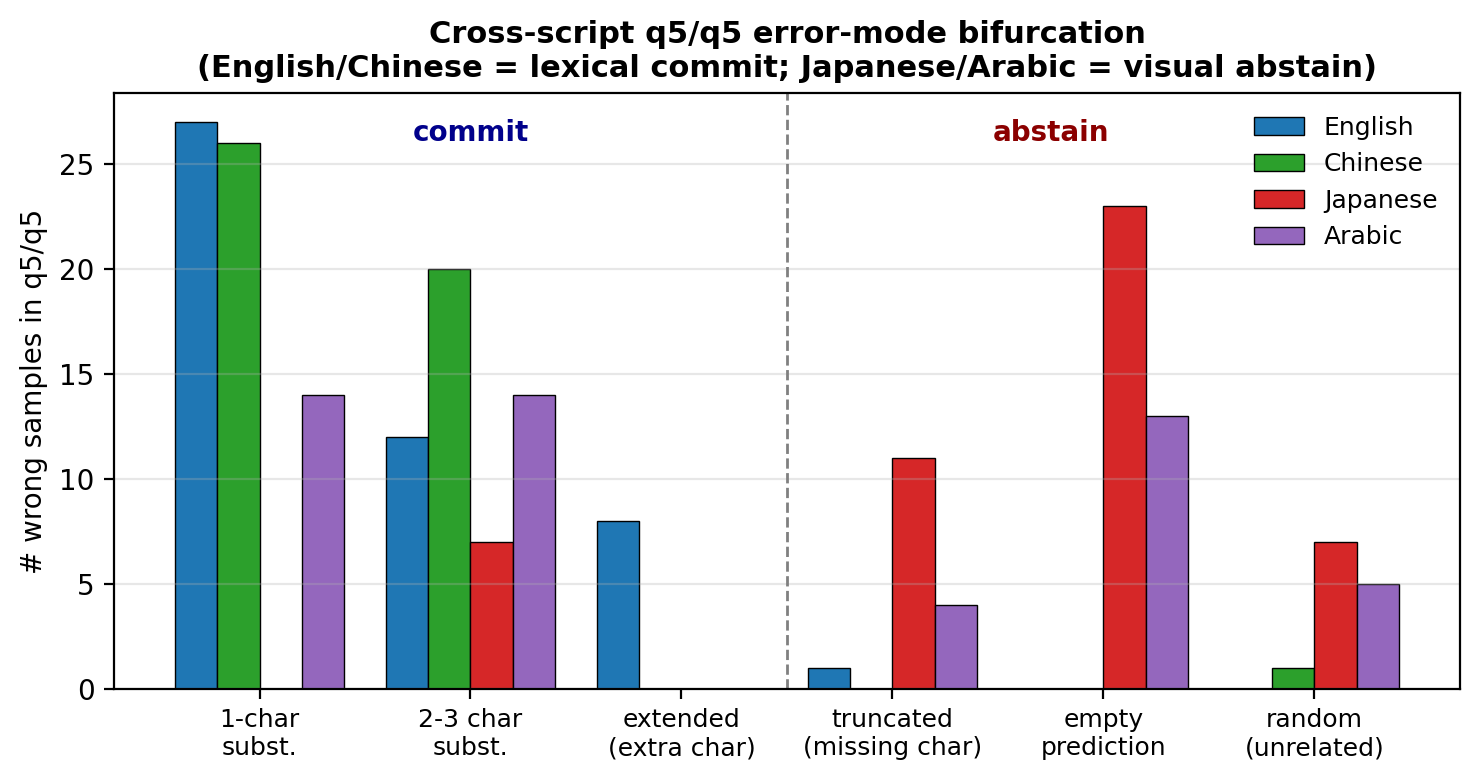}
\caption{Cross-script q5/q5 error-mode bifurcation. Stack of fifty
sampled wrong predictions per script, categorised by edit distance
from the ground truth. Both regimes are consistent with the
lexical-prior account; relative weight depends on training-corpus
lexical density.}
\label{fig:xlang-supp}
\end{figure*}

This appendix expands the cross-script analysis of
Section~\ref{sec:exp-mechanism}. Figure~\ref{fig:xlang-supp} quantifies
the two error regimes; the galleries that follow show representative
q5/q5 wrong predictions per script.

Of $50$ sampled q5/q5 wrong predictions per script, English and
Chinese fail by lexical \emph{commit} (substitution toward a familiar
token, $94$\% and $92$\% of failures respectively); Japanese and Arabic
fail by visual \emph{abstain} (empty or truncated predictions, $68$\%
and $34$\% respectively).

\begin{figure}[h]
\centering
\includegraphics[width=\linewidth]{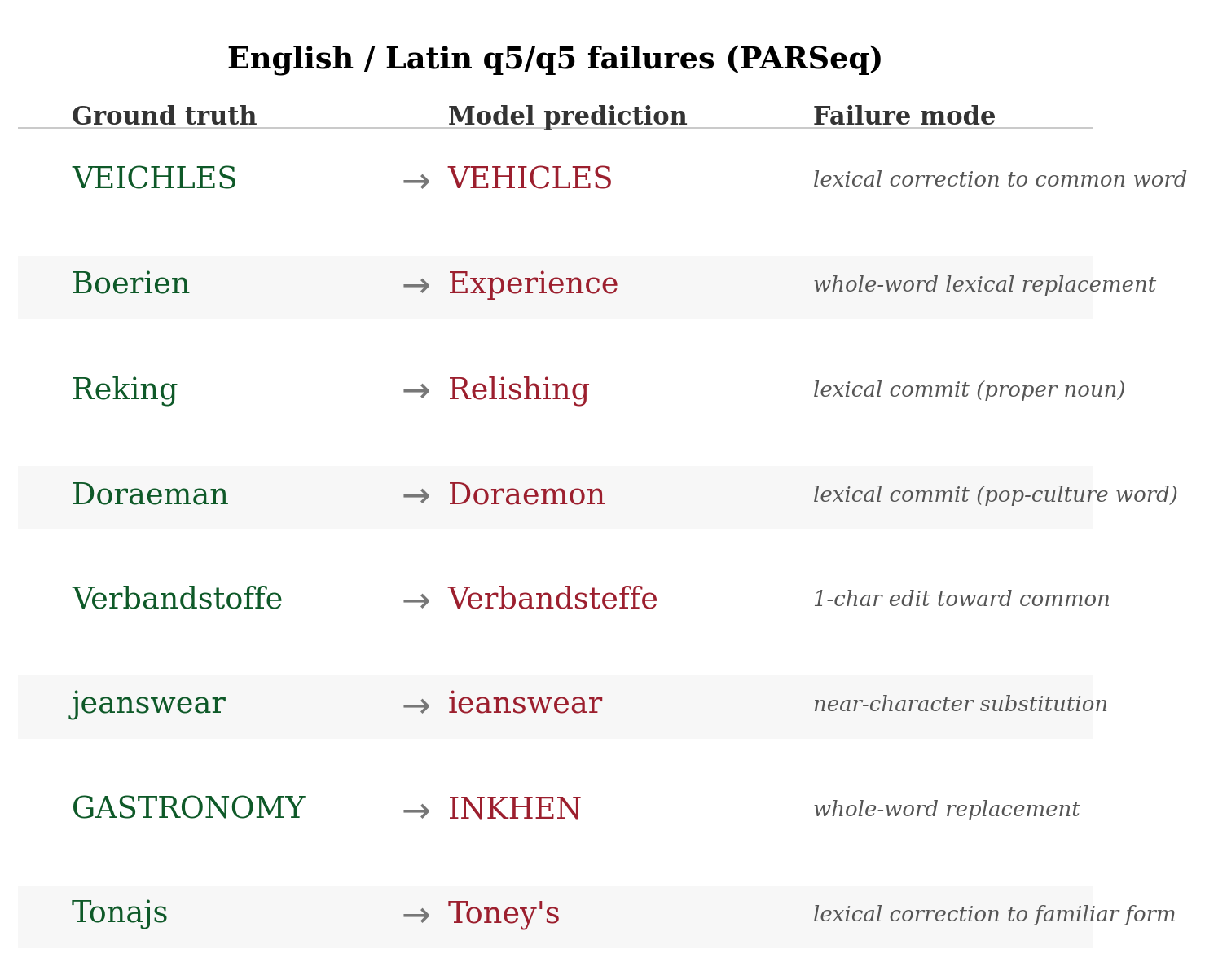}
\caption{\textbf{English / Latin q5/q5 failures (PARSeq).} All shown
examples are \emph{commit}-type errors: the AR decoder snaps to the
nearest familiar English word (\emph{VEHICLES}, \emph{Experience},
\emph{Relishing}, \emph{Doraemon}) or applies a 1-character edit
toward a more common form. Failures are systematic, not random
character noise.}
\label{fig:xlang-en-supp}
\end{figure}

\begin{figure}[h]
\centering
\includegraphics[width=\linewidth]{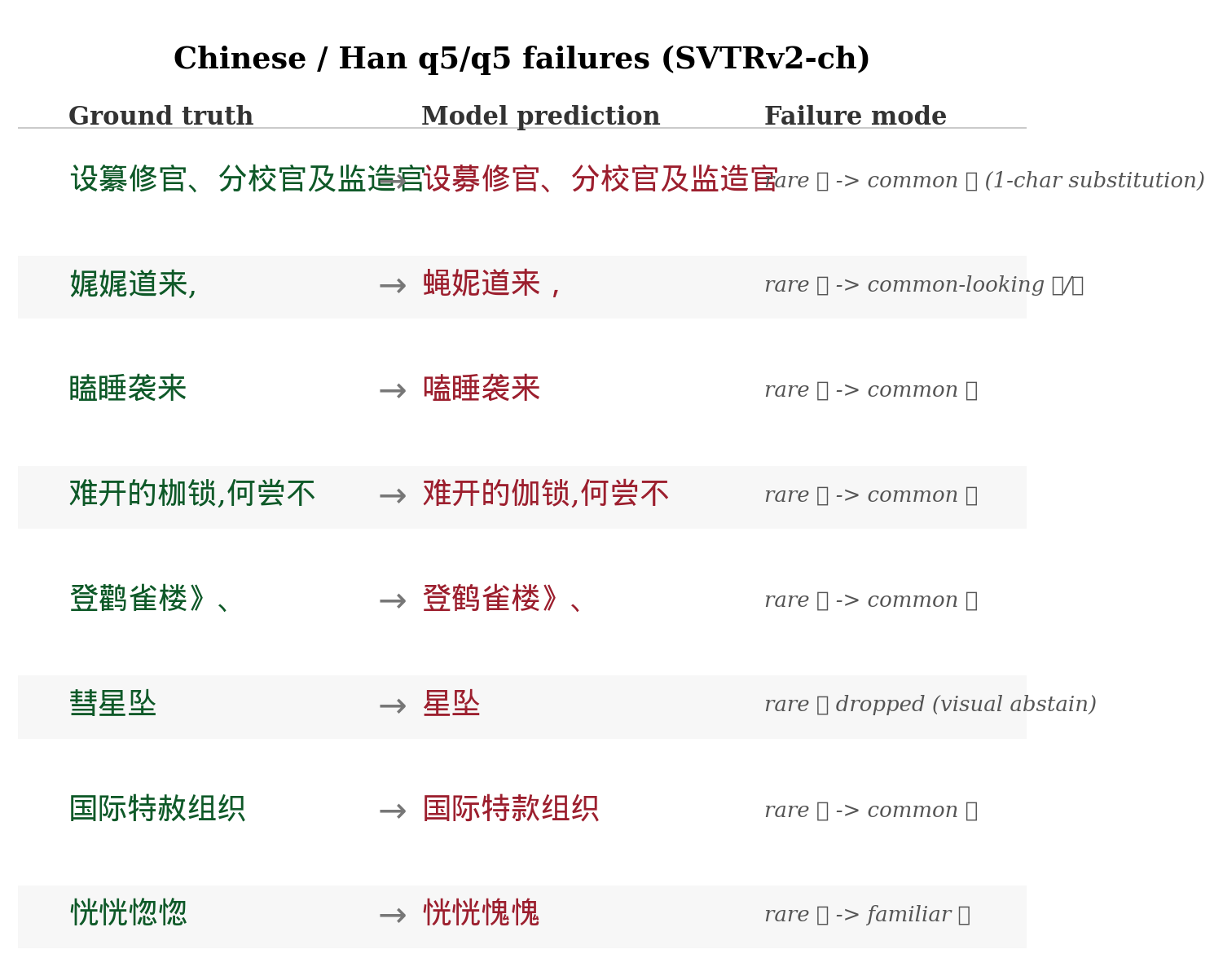}
\caption{\textbf{Chinese / Han q5/q5 failures (SVTRv2-ch).} The
failure mode is per-character substitution toward a more common
visually similar character; rare characters in the ground truth
are systematically replaced by their high-frequency lookalikes
(see the rightmost column). The model commits to the most
lexically familiar glyph in the local neighbourhood.}
\label{fig:xlang-zh-supp}
\end{figure}

\begin{figure}[h]
\centering
\includegraphics[width=\linewidth]{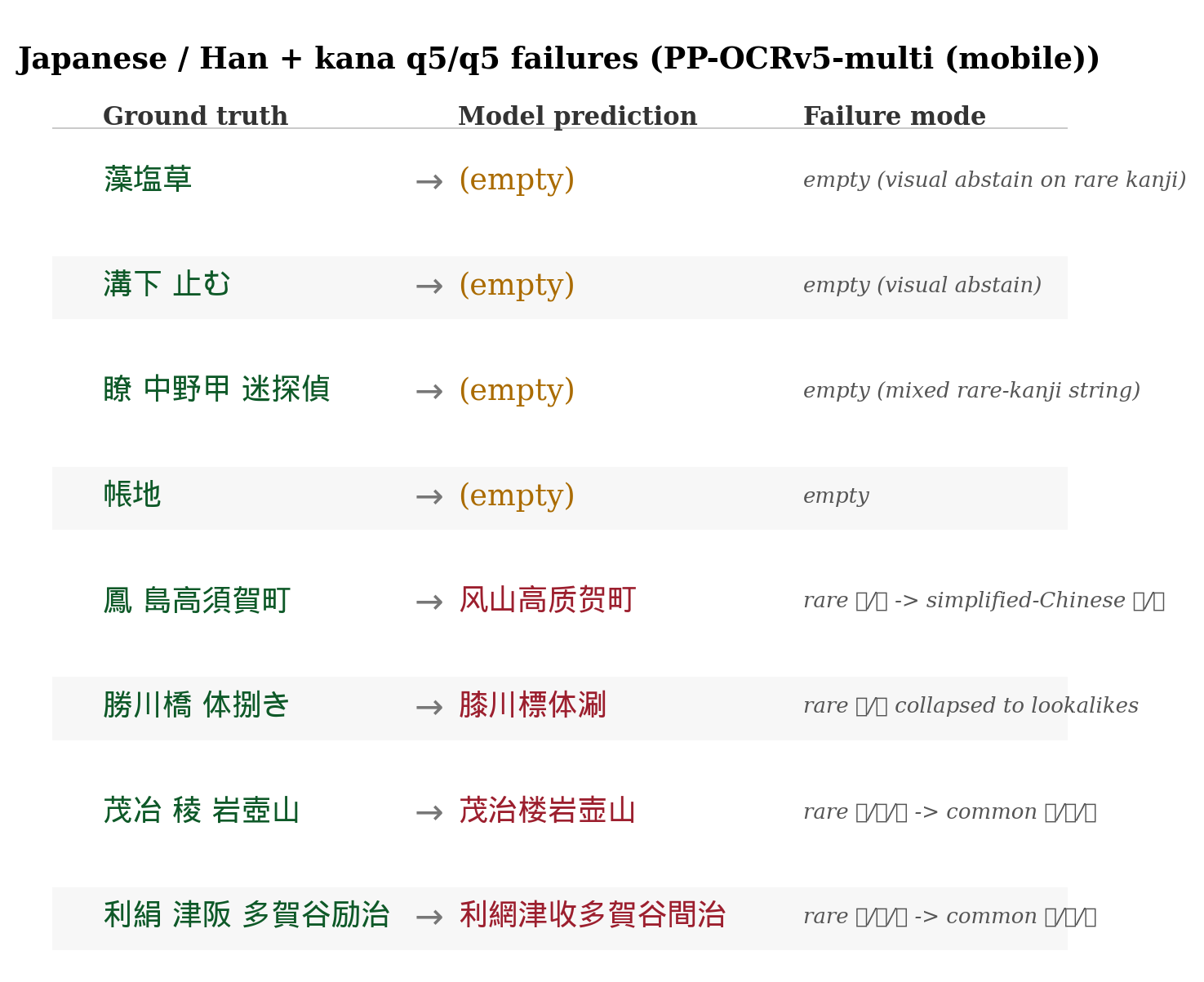}
\caption{\textbf{Japanese / Han + kana q5/q5 failures (PP-OCRv5-multi
mobile).} The dominant failure mode is \emph{abstain}: the recogniser
returns an empty prediction on rare-kanji strings (top four rows).
When it does commit, rare Japanese kanji are replaced by visually
similar simplified-Chinese characters (bottom four rows; see
rightmost column for per-row annotations).}
\label{fig:xlang-ja-supp}
\end{figure}

\begin{figure}[h]
\centering
\includegraphics[width=\linewidth]{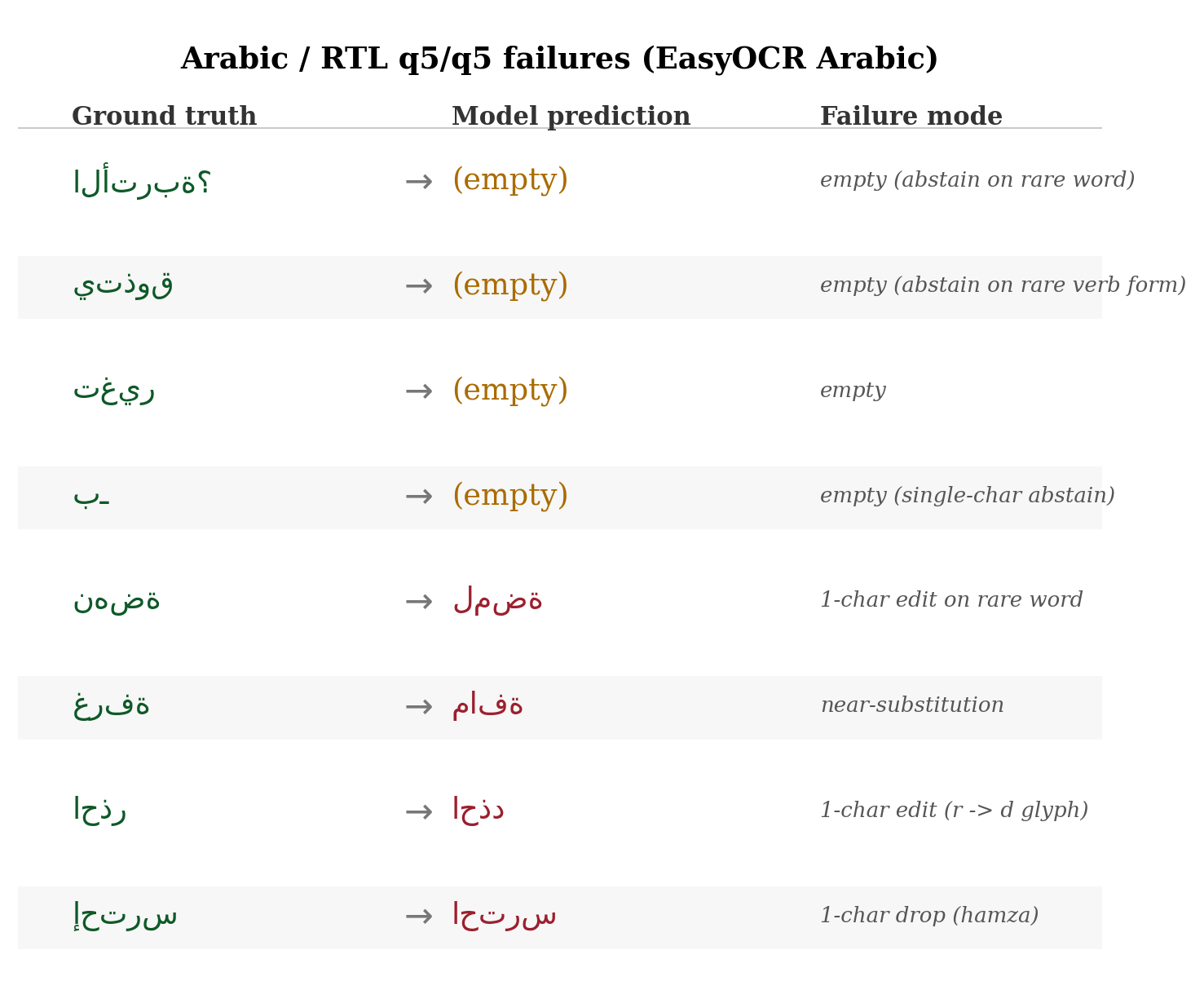}
\caption{\textbf{Arabic / RTL q5/q5 failures (EasyOCR Arabic).}
Mixed regime: roughly one third of failures are pure \emph{abstain}
(empty predictions on rare verb forms and isolated function words);
the remainder are single-character edits, near-substitutions, or
hamza drops (see annotations in rightmost column). The smaller
Arabic alphabet ($28$ root letters) produces fewer ``rare-character''
partitions, which we suggest is why the gap magnitude is smaller in
Arabic ($-4.9$\,pt) than in Han-script benchmarks ($-10$ to $-18$\,pt).}
\label{fig:xlang-ar-supp}
\end{figure}

The four galleries below display representative q5/q5 wrong
predictions for each writing system. Each sample comes from the
language's strongest recogniser on the relevant benchmark, and the
right-most column annotates the failure mode. The English and Chinese
galleries (Figures~\ref{fig:xlang-en-supp} and~\ref{fig:xlang-zh-supp})
are dominated by \emph{commit} failures; the Japanese and Arabic
galleries (Figures~\ref{fig:xlang-ja-supp} and~\ref{fig:xlang-ar-supp})
are dominated by \emph{abstain} failures.

\end{document}